%% file: ToArxiv.tex
\documentclass[letterpaper]{article} 
\usepackage{aaai2026}  
\usepackage{times}  
\usepackage{helvet}  
\usepackage{courier}  
\usepackage[hyphens]{url}  
\usepackage{graphicx} 
\usepackage{natbib}  
\usepackage{caption} 
\usepackage{float}
\usepackage{listings}
\DeclareCaptionStyle{ruled}{labelfont=normalfont,labelsep=colon,strut=off} 
\floatstyle{ruled}
\newfloat{listing}{tb}{lst}{}
\floatname{listing}{Listing}

\usepackage{latexsym}
\usepackage{url}

\usepackage{xspace}
\usepackage{graphicx}
\usepackage{multirow}
\usepackage{makecell}
\usepackage{lscape}
\usepackage{bm}
\usepackage{algorithm}
\usepackage{algpseudocode}
\usepackage{framed}
\usepackage{color}
\usepackage{xcolor}
\usepackage{booktabs,colortbl}
\usepackage{subfigure}
\usepackage{multirow}

\usepackage{amssymb}
\usepackage{amsmath}
\usepackage{multirow}
\usepackage{makecell}
\usepackage{tabularx}
\usepackage{booktabs}
\usepackage{utfsym}
\usepackage{pifont}
\usepackage{mathrsfs}

\definecolor{lightgrayv}{HTML}{F4F3F8} 
\definecolor{redv}{HTML}{C00000}
\definecolor{bluev}{HTML}{1F4E79} 
\definecolor{grayv}{HTML}{707070} %

\newcommand{\eg}{\emph{e.g.,}\xspace}

\newcommand{\wrt}{\emph{w.r.t}\xspace}
\newcommand{\baby}{\textsc{RetSimd}\xspace}
\title{Enhancing Multimodal Misinformation Detection by Replaying the Whole Story from Image Modality Perspective}
\author{
    Bing Wang\textsuperscript{\rm 1,2}, 
    Ximing Li\textsuperscript{\rm 1,2}\thanks{Ximing Li and Shengsheng Wang are corresponding authors.}
    Yanjun Wang\textsuperscript{\rm 2,3}, 
    Changchun Li\textsuperscript{\rm 1,2},  
    Lin Yuanbo Wu\textsuperscript{\rm 4},
    Buyu Wang\textsuperscript{\rm 5},
    Shengsheng Wang\textsuperscript{\rm 1,2*}
    \\
}
\affiliations{
    \textsuperscript{\rm 1}  College of Computer Science and Technology, Jilin University, Changchun, Jilin, China \\
    \textsuperscript{\rm 2}  Key Laboratory of Symbolic Computation and Knowledge Engineering of the MoE, Jilin University, Changchun, Jilin, China \\
    \textsuperscript{\rm 3}  College of Software, Jilin University, China
    \textsuperscript{\rm 4}  School of Engineering, University of Warwick, Coventry, UK \\
    \textsuperscript{\rm 5}  College of Computer and Information Engineering, Inner Mongolia Agricultural University, Hohhot, Inner Mongolia, China \\
    \{wangbing1416,liximing86,changchunli93\}@gmail.com, wss@jlu.edu.cn
}

\usepackage{bibentry}

\begin{document}

\maketitle

\begin{abstract}
\input{S_Abstract}
\end{abstract}

\begin{links}
    \link{Code}{https://github.com/wangbing1416/RETSIMD}
\end{links}

\input{S_Introduction}

\input{S_Analysis}

\input{S_Method}

\input{S_Experiment}

\input{S_Relatedworks}

\input{S_Conclusion}

\section*{Acknowledgement}

We acknowledge support for this project from the National Science and Technology Major Project (No. 2021ZD0112500), the National Natural Science Foundation of China (No. 62276113), and the China Postdoctoral Science Foundation (No. 2022M721321).

\bibliography{reference}

\input{S_Appendix}

\end{document}

%% file: S_Abstract.tex
Multimodal Misinformation Detection (MMD) refers to the task of detecting social media posts involving misinformation, where the post often contains text and image modalities. However, by observing the MMD posts, we hold that the text modality may be much more informative than the image modality because the text generally describes the whole event/story of the current post but the image often presents partial scenes only. Our preliminary empirical results indicate that the image modality exactly contributes less to MMD. Upon this idea, we propose a new MMD method named \baby. Specifically, we suppose that each text can be divided into several segments, and each text segment describes a partial scene that can be presented by an image. Accordingly, we split the text into a sequence of segments, and feed these segments into a pre-trained text-to-image generator to augment a sequence of images. We further incorporate two auxiliary objectives concerning text-image and image-label mutual information, and further post-train the generator over an auxiliary text-to-image generation benchmark dataset. Additionally, we propose a graph structure by defining three heuristic relationships between images, and use a graph neural network to generate the fused features. Extensive empirical results validate the effectiveness of \baby.

%% file: S_Introduction.tex
\section{Introduction}

Over the past decade, social media platforms, \eg Twitter and Weibo, have become the priority medium for delivering a plethora of messages and information to people worldwide. Unfortunately, they also promote the spread of misinformation about social topics, especially for hot events and stories, resulting in various negative effects \citep{vosoughi2018spread,scheufele2019science}. 
To alleviate such impacts, detecting misinformation timely becomes a primary demand in many real-world platforms, giving birth to the prevalent research topic, namely \textbf{M}isinformation \textbf{D}etection (\textbf{MD}) \citep{zhang2021mining,zhu2022generalizing,wang2024why}, in the data mining community. 

\begin{figure}[t]
  \centering
  \includegraphics[scale=0.27]{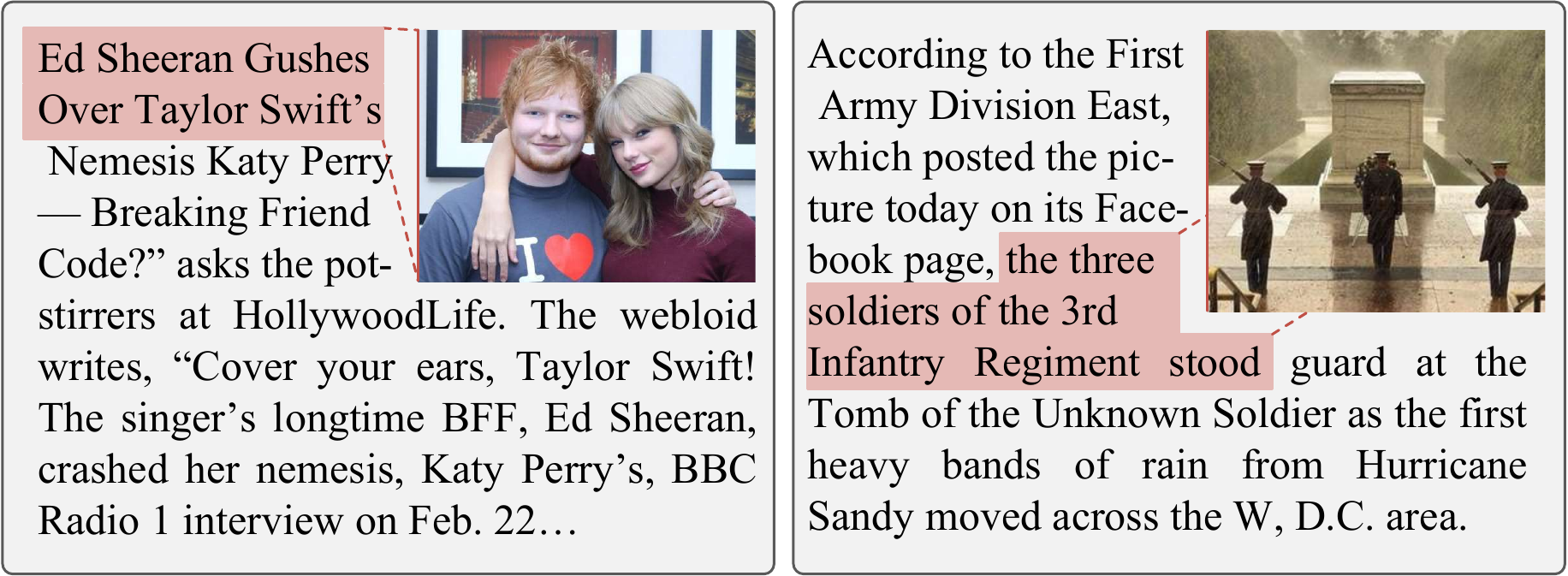}
  \caption{Two real-world MMD post examples. Each text describes the whole story of the current post but the image may only present a partial scene corresponding to the text segment marked in red.}
  \label{example}
\end{figure}

To our knowledge, the previous studies regard MD as a binary classification problem, and they mainly train neural-based models, which can predict any future post content whether it describes a real or fake event/story. Nowadays, many posts from social media are accompanied by images, so one can formulate such posts as multimodal samples with text and image modalities. Inspired by the cognition that multiple modalities can provide various views to express the post content, the community has paid much attention to developing \textbf{M}ultimodal \textbf{M}isinformation \textbf{D}etection (\textbf{MMD}) methods \citep{jin2017multimodal,chen2022cross,wang2024harmfully}. Generally, the prevalent ideas of MMD methods include fusing text and image modalities to achieve more discriminative features \citep{ying2023bootstrapping,wu2023see}, and learning the semantic inconsistency between text and image modalities as auxiliary cues for verifying misinformation \citep{fung2021infosurgeon,chen2022cross}.

A potential assumption is that, in the MMD task, the text and image modalities are equally important \citep{hu2023causal,wang2024fake,ma2024event}. However, by observing the MMD posts, we hold that the text modality may be much more informative than the image modality because the text generally describes the whole event/story of the current post but the image often presents partial scenes only (see examples in Fig.~\ref{example}). Accordingly, we argue that the text modality must be more important than the image modality in MMD. To verify this viewpoint, we conduct extensive preliminary experiments to evaluate the contributions of text and image modalities by predicting ablative variants of SOTA MMD methods. The results validate that, for each MMD method, the performance gap between the full variant and the variants with text modalities is significantly and consistently lower than the one between the full variant and the variants with image modalities, empirically proving the argument.

Based on these observations, a natural way of promoting MMD is, for each MMD sample, to augment a sequence of images that can also replay the whole story of the post, instead of employing the image that often presents partial scenes only. Upon this idea, we propose a new MMD method named \textbf{\baby}. 
Specifically, we suppose that each text can be divided into several segments, and each text segment describes a partial scene that can be presented by an image. Accordingly, we split the text into a sequence of segments, and feed these segments into a pre-trained text-to-image generator to augment a sequence of images. To guarantee higher-quality augmented images, we incorporate two auxiliary objectives concerning text-image and image-label mutual information, and further post-train the generator over an auxiliary text-to-image generation benchmark dataset. With the augmented images, we propose a graph-based encoder to integrate their features. Specifically, we construct a graph structure by defining three heuristic relationships between images, and use a graph neural network to encode the graph to obtain the fused features.

We conduct extensive experiments across three benchmark MMD datasets and compare five SOTA MMD methods. The experimental results indicate that \baby can consistently improve the performance of baseline models, which proves the effectiveness of \baby. Meanwhile, we also report quantitative results demonstrating that \baby effectively improves the contributions of the image modality to MMD.

\begin{figure*}[t]
  \centering
  \includegraphics[scale=0.150]{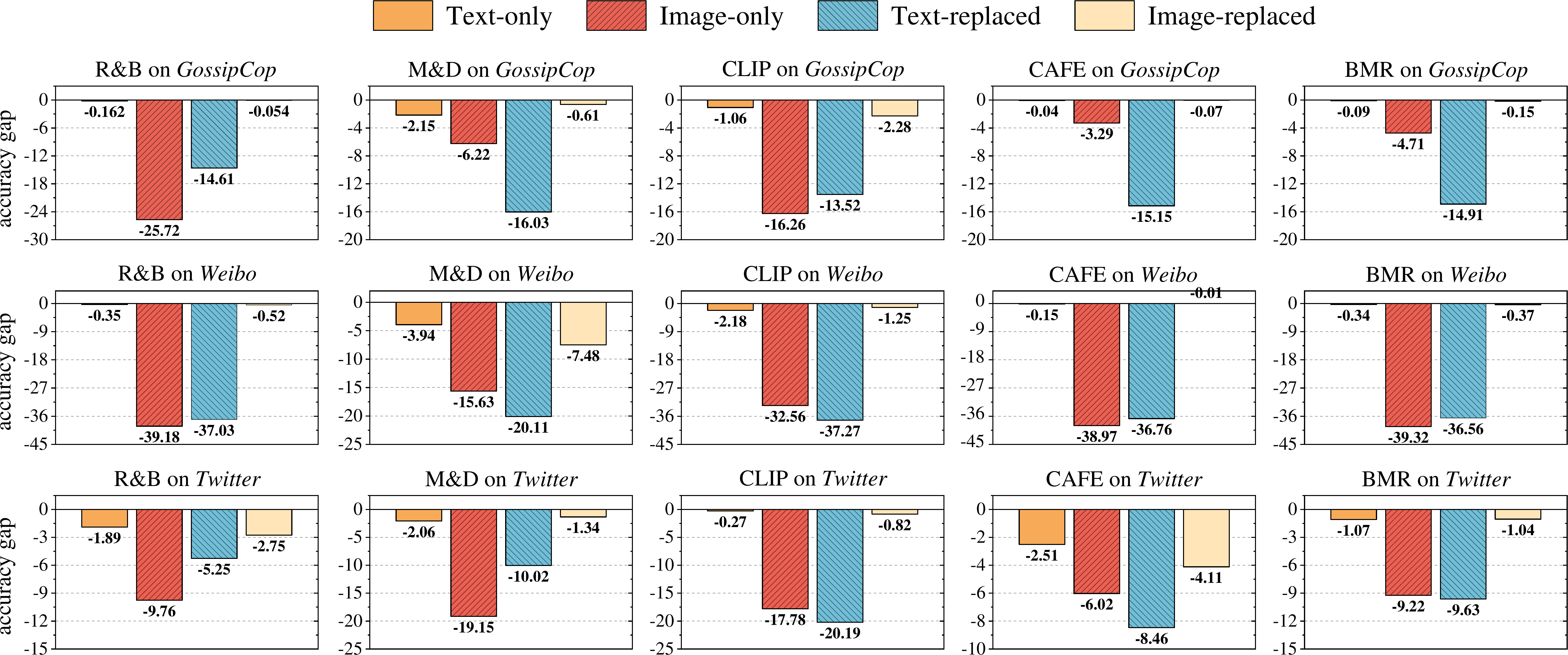}
  \caption{Preliminary empirical evaluations on accuracy gaps between four ablative variants and the full variant.}
  \label{evaluation_acc}
\end{figure*}

\begin{figure*}[t]
  \centering
  \includegraphics[scale=0.150]{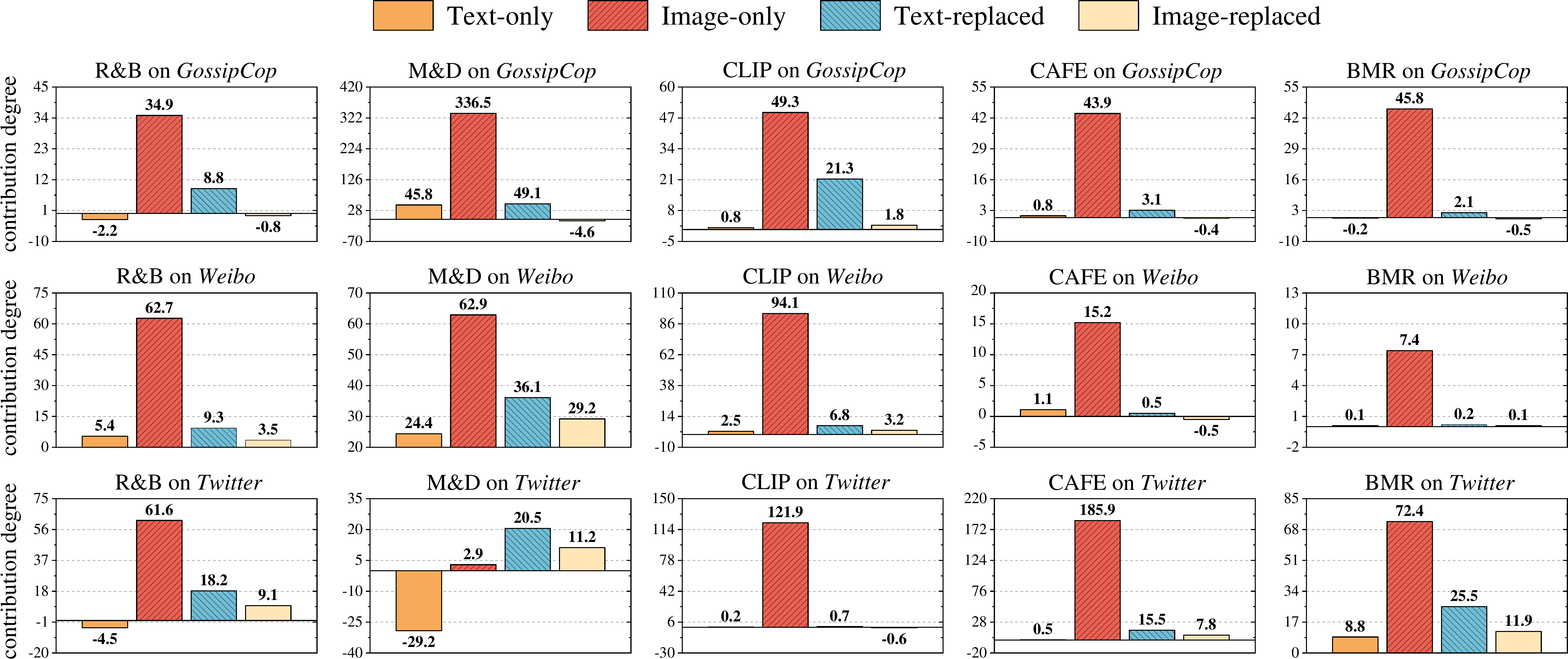}
  \caption{Preliminary empirical evaluations on contribution degrees of four ablative variants.}
  \label{evaluation_ig}
\end{figure*}

Our contributions are the following three-fold:
\begin{itemize}
    \item We empirically validate that the text modality contributes more to MMD models than the image modality.
    \item We propose a new MMD method \baby by generating a sequence of augmented images that can also replay the whole story of the text.
    \item Extensive experiments are conducted to validate the effectiveness of \baby.
\end{itemize}

%% file: S_Analysis.tex
\section{Preliminary Empirical Evaluations on Modality Contributions} \label{sec:analysis}

In this evaluation, we employ five SOTA MMD methods including ResNet + BERT \textbf{(R\&B)} \citep{he2016deep,devlin2019bert}, MAE + DeBERTaV3 \textbf{(M\&D)} \citep{he2022masked,he2023debertav3}, \textbf{CLIP} \citep{radford2021learning}, \textbf{CAFE} \citep{chen2022cross}, and \textbf{BMR} \citep{ying2023bootstrapping}. To evaluate the modality contributions in MMD, for each MMD method, we compare the classification accuracy scores between the following ablative variants:
\begin{itemize}
    \item \textbf{Full variant} trains using the original text-image pairs.
    \item \textbf{Text-only variant} removes the images by setting them to a completely white image, leaving only the text.
    \item \textbf{Image-only variant} removes the texts by setting them to the same-length \texttt{[PAD]} tokens, leaving only the images.
    \item \textbf{Text-replaced variant} replaces the texts in the original image-text pairs with the ones in other samples.
    \item \textbf{Image-replaced variant} replaces the images in the original image-text pairs with the ones in other samples.
\end{itemize}
We employ three benchmark MMD datasets including \textit{GossipCop} \citep{shu2020fakenewsnet}, \textit{Weibo} \citep{jin2017multimodal}, and \textit{Twitter} \citep{boididou2018detection}. The descriptions of benchmark datasets, compared methods, and implementation details are presented in the Supplementary Material.

\subsection{Results and Discussion}

Because the full variants perform the best in all the cases, we show the accuracy gap between the full variants and the other four ablative variants in Fig.~\ref{evaluation_acc}. First, we can observe that the gaps of text-only/image-replaced variants are lower than those of image-only/text-replaced variants by a large margin and, in some cases, the image-only/text-replaced variants are even ineffective to some extent. These results directly validate the argument that the text modality contribution is much more than the image modality to MMD. Further, we can see that the accuracy gaps of text-only and image-replaced variants are consistently insignificant in most cases. That is because the text can provide sufficient information to describe the content, further supporting the argument.

\subsection{Quantifying Modality Contributions}

In the above section, we conclude that the text modality contributes more to MMD than the image modality. Meanwhile, in Fig.~\ref{example}, we analyze that the different contributions of the two modalities are attributed to the fact that text always describes the whole story but images only depict partial scenes. 
This inspires us to investigate the modality contributions from the information theory perspective. Accordingly, we are motivated by the information gain \citep{quinlan1986induction,donoho1995de}, and design a quantitative metric named \textbf{contribution degree} for our experiments.
Formally, given an image-text pair $(\mathbf{x}^v, \mathbf{x}^t)$, we calculate its information entropy \wrt the veracity label $y$ as follows:
\begin{equation}
    \label{eq1}
     \mathcal{H}(y | \mathbf{x}^v, \mathbf{x}^t) = - \sum p(y | \mathbf{x}^v, \mathbf{x}^t) \log p(y | \mathbf{x}^v, \mathbf{x}^t).
\end{equation}
Upon it, we can calculate the information gains carried by image and text modalities as follows:
\begin{equation}
    \label{eq2}
    \begin{aligned}
        \mathscr{G}(y, \mathbf{x}^v) = \mathcal{H}(y | \varnothing , \mathbf{x}^t) - \mathcal{H}(y | \mathbf{x}^v, \mathbf{x}^t), \\
        \mathscr{G}(y, \mathbf{x}^t) = \mathcal{H}(y | \mathbf{x}^v , \varnothing) - \mathcal{H}(y | \mathbf{x}^v, \mathbf{x}^t).
    \end{aligned}
\end{equation}
In addition to the information gains brought by such two variants, we also investigate the information gains of text-replaced and image-replaced variants $\mathscr{G}(y, \mathbf{\hat x}^t)$ and $\mathscr{G}(y, \mathbf{\hat x}^v)$, the experimental results are reported in Fig.~\ref{evaluation_ig}.

Generally, the information gains brought by the image-only/text-replaced variants are larger than those of text-only/image-replaced variants. The observation indicates that text contributes the most substantial information gain for veracity predictions, while images contribute the least, or even negative gain. Therefore, these findings quantitatively demonstrate the greater importance of the text modality over the image modality, and provide support for our hypothesis that the contribution gap exists between the two modalities. Inspired by the evidence, we aim to generate more images that depict the whole stories presented in the text modality to bridge the information gap between the two modalities.

%% file: S_Method.tex
\section{Our Proposed Method}

\begin{figure*}[t]
  \centering
  \includegraphics[scale=0.685]{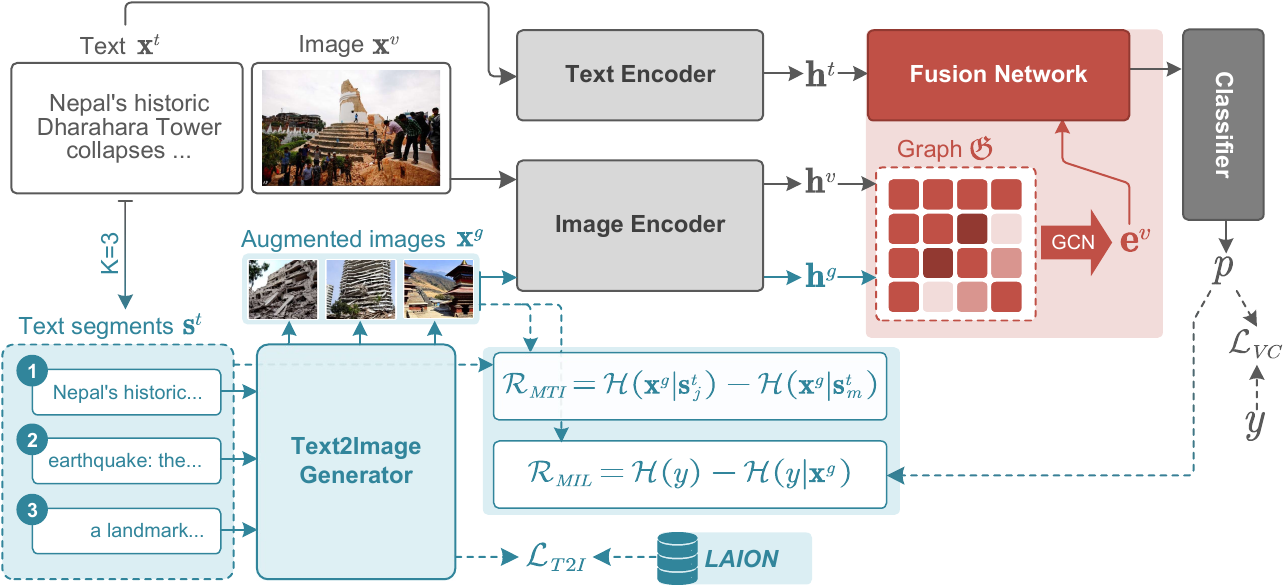}
  \caption{Overall framework of \baby. We split $\mathbf{x}^t$ into $K$ segments, and use a text-to-image generator to generate corresponding images $\mathbf{x}^g$. Then, we design a multimodal fusion network to integrate these features and predict the veracity label.}
  \label{framework}
\end{figure*}

In this section, we introduce the task formulation of MMD and our proposed MMD framework \baby, which aims to improve the contribution of the image modality to MMD.

\vspace{2pt} \noindent
\textbf{Task formulation.}
Typically, an MMD dataset is formulated as $\mathcal{D} = \{\mathcal{X}_i, y_i\}_{i=1}^N$, where $\mathcal{X}_i = (\mathbf{x}_i^t, \mathbf{x}_i^v)$ denotes the text-image pair, and $y_i \in \{0, 1\}$ is its ground-truth veracity label, \eg real and fake. Accordingly, the general goal of MMD is to train a detector $\mathcal{F}_{\boldsymbol{\theta}}(\cdot)$, which can predict the veracity labels of unseen text-image pairs. 

\subsection{Model Overview}

In Sec.~\ref{sec:analysis}, we empirically observe that the text modality contribution is much more than the image modality to MMD.
And the experimental results support the argument that the text may completely describe the whole story, but the image can only partially present one scene in the story. Accordingly, we attempt to improve the contribution of the image modality by generating a sequence of augmented images \wrt the text modality. Based on this idea, we propose a new MMD framework \baby, which alternately optimizes the misinformation detector and a text-to-image generator, which generates images to enhance the image modality, and then we design a fusion model to integrate these images by mining their potential relationships.
Specifically, \baby generally includes four modules: \textbf{feature encoders}, \textbf{text-to-image generator}, \textbf{multimodal fusion network}, and \textbf{veracity classifier}. For clarity, the overall framework of \baby is depicted in Fig.~\ref{framework}.

\vspace{2pt} \noindent
\textbf{Feature encoders.} 
Given text $\mathbf{x}_i^t$ and an image $\mathbf{x}_i^v$, the feature encoders utilize the text encoder and image encoder to extract their shared semantic features $\mathbf{h}_i^t$ and $\mathbf{h}_i^v$.
Specifically, we first leverage a pre-trained language model BERT \citep{devlin2019bert} and a ResNet34 model \citep{he2016deep} pre-trained by ImageNet \citep{krizhevsky2012imagenet} to obtain the hidden text and image features $\mathbf{z}_i^t = \mathcal{F}_{\boldsymbol{\theta}^t}(\mathbf{x}_i^t)$ and $\mathbf{z}_i^v = \mathcal{F}_{\boldsymbol{\theta}^v}(\mathbf{x}_i^v)$, respectively. 
And then, to align these features into a shared space, we use linear layers to obtain two shared features $\mathbf{h}_i^t = \mathbf{z}_i^t \mathbf{W}_a^t$ and $\mathbf{h}_i^v = \mathbf{z}_i^v \mathbf{W}_a^v$.

\vspace{2pt} \noindent
\textbf{Text-to-image generator.} 
To improve the contribution of the image modality to MMD, we design a text-to-image generator $\mathcal{G}_{\boldsymbol{\phi}}(\cdot)$ to generate a sequence of augmented images corresponding to the text.
Specifically, given a piece of text $\mathbf{x}_i^t$, we aim for the augmented images to completely express its described story. Therefore, we divide $\mathbf{x}_i^t$ into $K$ segments with a fixed-number sliding window\footnote{We evaluate multiple text segmentation strategies in the Appendix, and the fixed-number sliding has the best performance.} as $\{\mathbf{s}_{ij}^t\}_{j=1}^K$. Then, we feed these text segments into the generator to generate images $\mathbf{x}_{ij}^g = \mathcal{G}_{\boldsymbol{\phi}}(\mathbf{s}_{ij}^t)$.

Meanwhile, to ensure that the augmented images $\mathbf{x}_{ij}^g$ contain sufficient information and can fully express the information in the text $\mathbf{s}_{ij}^t$, we design two information theory-based regularizations $\mathcal{R}_{MTI}$ and $\mathcal{R}_{MIL}$ to tune the generator, which learn the mutual information between the text and its generated images, and between the generated images and the veracity labels, respectively.
We also utilize prevalent text-to-image datasets, \eg \textit{LAION-2B} \citep{schuhmann2022laion}, to post-train the generator by an objective $\mathcal{L}_{T2I}$. In summary, the overall training objective of the text-to-image generator is as follows:
\begin{equation}
    \label{eq4}
    \mathop{\boldsymbol{\min}} \limits _{\boldsymbol{\phi}} \mathcal{L}_{GEN}(\boldsymbol{\phi}) = \mathcal{L}_{T2I} + \alpha_1 \mathcal{R}_{MTI} + \alpha_2 \mathcal{R}_{MIL},
\end{equation}
where $\alpha_1$ and $\alpha_2$ represent two trade-off parameters to balance multiple objectives.

\vspace{2pt} \noindent
\textbf{Multimodal fusion network.} 
Given augmented images $\{\mathbf{x}_{ij}^g\}_{j=1}^K$ generated by the text-to-image generator, the multimodal fusion network integrates their features with the original image $\mathbf{x}_i^v$ and the text $\mathbf{x}_i^t$. Specifically, to grasp the potential relationships among different images, we construct a graph structure $\mathfrak{G}$, where the nodes represent all the images, and the edges indicate our designed relationships between the image nodes. Upon this graph, we leverage a prevalent graph neural network \citep{kipf2017semi} to capture a fused image feature $\mathbf{e}_i^v = \mathcal{F}_{\boldsymbol{\theta}^g}(\mathbf{h}_i^v, \mathbf{h}_{ij}^g, \mathfrak{G})$, where $\mathbf{h}_{ij}^g = \mathcal{F}_{\boldsymbol{\theta}^v}(\mathbf{x}_{ij}^g) \mathbf{W}_a^v$ denotes the shared feature of the augmented images. Given the text feature $\mathbf{h}_i^t$ and the fused image feature $\mathbf{e}_i^v$, we use a cross-attention network to obtain the overall fused feature $\mathbf{e}_i = \mathcal{F}_{\boldsymbol{\theta}^c}(\mathbf{h}_i^t, \mathbf{e}_i^v)$.

\vspace{2pt} \noindent
\textbf{Veracity classifier.} 
Utilizing the fused feature $\mathbf{e}_i$, we feed it into a veracity classifier consisting of two linear layers and an activation function, to predict its veracity label $p_i = \mathbf{e}_i \mathbf{W}_v$. Accordingly, the training objective of the veracity classification is as follows:
\begin{equation}
    \label{eq5}
    \mathcal{L}_{VC} = \frac{1}{N} \sum \nolimits _{i=1}^N \ell_{CE}(p_i, y_i),
\end{equation}
where $\ell_{CE}(\cdot\ , \cdot)$ represents a cross-entropy loss function. Upon the aforementioned modules, the overall training objective of the misinformation detector is as follows:
\begin{equation}
    \label{eq6}
    \mathop{\boldsymbol{\min}} \limits _{\boldsymbol{\theta}} \mathcal{L}_{DET}(\boldsymbol{\theta}) = \mathcal{L}_{VC} + \beta \mathcal{R}_{CA},
\end{equation}
where $\beta$ is a hyper-parameter. In summary, the training of \baby is to alternately optimize Eq.~\eqref{eq4} and Eq.~\eqref{eq6} \wrt $\boldsymbol{\theta}$ and $\boldsymbol{\phi}$, respectively.
In the following sections, we describe the details of the text-to-image generator and the multimodal fusion network, respectively.

\subsection{Text-to-image Generator}

In general, we tune the text-to-image generator $\mathcal{G}_{\boldsymbol{\phi}}(\cdot)$ , which generates images $\{\mathbf{x}_{ij}^g\}_{j=1}^K$ with the text $\mathbf{x}_i^t$ to make the image modality contributing. Specifically, we use the stable diffusion model \citep{rombach2022high} pre-trained across text-image pairs as the basic generator. Given $\mathbf{x}_i^t$, we split it into $K$ equal-length segments $\{\mathbf{s}_{ij}^t\}_{j=1}^K$, and feed them into the generator to generate augmented images $\{\mathbf{x}_{ij}^g\}_{j=1}^K$. 
Upon the generated ones, we propose the following two mutual information objectives to further tune the generator.

\vspace{2pt} \noindent
\textbf{Text-image mutual information.}
Typically, the augmented images should be semantically consistent to their conditioned text descriptions. Therefore, we design an objective to constrain the mutual information between the text $\mathbf{s}_{ij}^t$ and its corresponding augmented image $\mathbf{x}_{ij}^g$. 
Previous mutual information based method \citep{zhang2021cross} achieve this via contrastive losses among different texts and images. In fact, in our method, different segments in the same text naturally form positive and negative samples. Formally, our text-image mutual information objective is as follows:
\begin{equation}
    \label{eq7}
    \mathcal{R}_{MTI} = \frac{1}{NK} \sum \nolimits _{i=1}^N \sum \nolimits _{j=1}^K \mathfrak{R} \big( \mathbf{s}_{ij}^t, \mathbf{x}_{ij}^g \big),
\end{equation}
where,
\begin{equation}
    \label{eq8}
    \small
    \mathfrak{R} \big( \mathbf{s}_{ij}^t, \mathbf{x}_{ij}^g \big) = \frac{1}{K - 1} \sum \nolimits _{m \neq j}
    \mathcal{H} \big( \mathbf{x}_{ij}^g | \mathbf{s}_{ij}^t \big) - \boldsymbol{\xi}_{jm} \mathcal{H} \big( \mathbf{x}_{ij}^g | \mathbf{s}_{im}^t \big),
\end{equation}
where $\mathcal{H}( \mathbf{x}_{ij}^g | \cdot)$ indicates the entropy of $\mathbf{x}_{ij}^g$ conditioned on different text segments. With this objective, we aim for $\mathbf{s}_{ij}^t$ to bring more information about $\mathbf{x}_{ij}^g$ than $\mathbf{s}_{im}^t$.
Additionally, $\boldsymbol{\xi}_{jm}$ denotes an adaptive weight. Since there is a temporal relationship between different text segments, the semantics of segments with close spatial distances are more relevant. Therefore, when $m$ is closer to $j$, the weight $\boldsymbol{\xi}_{jm}$ is smaller.

\vspace{2pt} \noindent
\textbf{Image-label mutual information.}
The goal of our augmented images is to make the image modality more informative regarding the veracity labels.
Therefore, we need to propose a objective to constrain the generator to achieve this goal. Accordingly, our image-label mutual information objective is formulated as follows:
\begin{align}
    \label{eq9}
    \mathcal{R}_{MIL} & = \frac{1}{N} \sum \nolimits _{i=1}^N \mathscr{G} \left( y_i | \{\mathbf{x}_{ij}^g\}_{j=1}^K \right) \nonumber \\ 
    & = \frac{1}{N} \sum \nolimits _{i=1}^N \mathcal{H}(y_i) - \mathcal{H} \left( y_i | \{\mathbf{x}_{ij}^g\}_{j=1}^K \right).
\end{align}
This objective utilizes the information gain brought by the generated images to the veracity labels, and uses it to train the generator in an end-to-end scheme.

The success of the aforementioned objectives always depends on the zero-shot capability of the pre-trained generator \citep{clark2023text}. To maintain high-quality images generated by the generator, we introduce a text-to-image dataset $\overline{\mathcal{D}} = \{\overline{\mathbf{x}}_i, \overline{y}_i\}_{i=1}^{|\overline{\mathcal{D}}|}$, \eg \textit{LAION}, to continuously post-train the generator, where $(\overline{\mathbf{x}}_i, \overline{y}_i)$ is a gold text-image pair. The post-training objective is as follows:
\begin{equation}
    \label{eq10}
    \mathcal{L}_{T2I} = \frac{1}{|\overline{\mathcal{D}}|} \sum \nolimits _{i=1}^{|\overline{\mathcal{D}}|} \ell_{MSE} \big( \mathcal{G}_{\boldsymbol{\phi}}(\overline{\mathbf{x}}_i ), \overline{y}_i \big),
\end{equation}
where $\ell_{MSE}(\cdot, \cdot)$ means the mean squared error function. Upon the objectives, we optimize the generator with Eq.~\eqref{eq4}.

\subsection{Multimodal Fusion Network}

Generally, the multimodal fusion network aims to integrate the features of the augmented images $\{\mathbf{x}_{ij}^g\}_{j=1}^K$ and the original text-image pair $(\mathbf{x}_i^t, \mathbf{x}_i^v)$. Specifically, by using the feature encoders, we can first obtain the hidden features $\{\mathbf{h}_{ij}^g\}_{j=1}^K$ and $\mathbf{h}_i^v$ of images.
To utilize these features more effectively, we capture the potential relationships among images, and construct a graph structure $\mathfrak{G}_i = (\boldsymbol{\Lambda}_i, \mathbf{H}_i)$, where $\boldsymbol{\Lambda}_i$ represents the adjacency matrix, the nodes are $K + 1$ images, the edges represent their relationships, and $\mathbf{H}_i = \{\mathbf{h}_i^v, \mathbf{h}_{i1}^g, \cdots, \mathbf{h}_{iK}^g\}$ denotes the node features.
Specifically, we design the following three heuristic relationships to construct the graph.

\vspace{2pt} \noindent
\textbf{Central relationship.}
We suggest that although the original images may only express the partial scene in an event, they always present the most central scene compared to images generated using different text segments. Therefore, based on this assumption, we directly connect all the augmented images with the original image to construct the initial graph.

\vspace{2pt} \noindent
\textbf{Temporal relationship.}
Our augmented images originate from the original text segments, and we expect that they depict the scenes of each text segment. In fact, a natural temporal relationship exists between the text segments, meaning that adjacent segments have a strong semantic correlation. Therefore, to learn this temporal relationship, we connect the augmented images according to the order of their corresponding text segments in the original text.

\begin{table*}[t]
\centering
\renewcommand\arraystretch{1.01}
  \caption{Experimental results of \baby across three prevalent MD datasets \textit{GossipCop}, \textit{Weibo}, and \textit{Twitter}. The results marked by * are statistically significant compared to its baseline models, satisfying p-value < 0.05.}
  \label{result}
  \footnotesize
  \setlength{\tabcolsep}{5pt}{
  \begin{tabular}{m{2.2cm}m{1.20cm}<{\centering}m{1.25cm}<{\centering}m{1.20cm}<{\centering}m{1.20cm}<{\centering}m{1.20cm}<{\centering}m{1.20cm}<{\centering}m{1.20cm}<{\centering}m{1.20cm}<{\centering}m{0.8cm}<{\centering}m{1.0cm}<{\centering}}
    \toprule
    \quad \quad Method & Accuracy & Macro F1 & P$_{\text{real}}$ & R$_{\text{real}}$ & F1$_{\text{real}}$ & P$_{\text{fake}}$ & R$_{\text{fake}}$ & F1$_{\text{fake}}$ & $\boldsymbol{\Delta}$ & $\mathscr{G}(y, \mathbf{x}^v)$  \\
    \hline
    
    \multicolumn{10}{c}{\textbf{Dataset: \textit{GossipCop}} \citep{shu2020fakenewsnet} } \\
    
    ResNet + BERT & 87.17{\color{grayv} \scriptsize $\pm$0.4} & 78.30{\color{grayv} \scriptsize $\pm$0.7} & 90.84{\small \color{grayv} \scriptsize $\pm$0.6} & 93.55{\color{grayv} \scriptsize $\pm$1.1} & 92.17{\color{grayv} \scriptsize $\pm$0.3} & 69.25{\color{grayv} \scriptsize $\pm$2.6} & {\textbf{60.41}}{\color{grayv} \scriptsize $\pm$3.5} & 64.43{\color{grayv} \scriptsize $\pm$1.4} & - & .0349 \\
    
    \rowcolor{lightgrayv} \quad + \textbf{\baby} & {\textbf{88.13}}{\color{grayv} \scriptsize $\pm$0.2}$^*$ & {\textbf{79.47}}{\color{grayv} \scriptsize $\pm$0.4}$^*$ & {\textbf{90.89}}{\color{grayv} \scriptsize $\pm$0.2} & {\textbf{94.79}}{\color{grayv} \scriptsize $\pm$0.3}$^*$ & {\textbf{92.80}}{\color{grayv} \scriptsize $\pm$0.1}$^*$ & {\textbf{73.38}}{\color{grayv} \scriptsize $\pm$0.9}$^*$ & 60.18{\color{grayv} \scriptsize $\pm$1.2} & {\textbf{66.13}}{\color{grayv} \scriptsize $\pm$0.7}$^*$ & {\textbf{1.21}} & {\textbf{.0301}} \\

    R\&B + SAFE & 87.14{\color{grayv} \scriptsize $\pm$0.6} & 78.74{\color{grayv} \scriptsize $\pm$0.3} & 91.35{\color{grayv} \scriptsize $\pm$0.6} & 92.88{\color{grayv} \scriptsize $\pm$1.6} & 92.10{\color{grayv} \scriptsize $\pm$0.4} & 68.22{\color{grayv} \scriptsize $\pm$3.6} & {\textbf{63.08}}{\color{grayv} \scriptsize $\pm$3.8} & 65.38{\color{grayv} \scriptsize $\pm$0.5} & - & .0325 \\
    
    \rowcolor{lightgrayv} \quad + \textbf{\baby} & {\textbf{88.30}}{\color{grayv} \scriptsize $\pm$1.1}$^*$ & {\textbf{79.79}}{\color{grayv} \scriptsize $\pm$0.7}$^*$ & {\textbf{91.02}}{\color{grayv} \scriptsize $\pm$0.7} & {\textbf{94.88}}{\color{grayv} \scriptsize $\pm$1.3}$^*$ & {\textbf{92.91}}{\color{grayv} \scriptsize $\pm$0.8}$^*$ & {\textbf{73.88}}{\color{grayv} \scriptsize $\pm$3.3}$^*$ & 61.73{\color{grayv} \scriptsize $\pm$3.1} & {\textbf{66.67}}{\color{grayv} \scriptsize $\pm$0.7}$^*$ & {\textbf{1.29}} & {\textbf{.0287}} \\

    R\&B + MCAN & 87.29{\color{grayv} \scriptsize $\pm$0.7} & 78.73{\color{grayv} \scriptsize $\pm$0.2} & 91.18{\color{grayv} \scriptsize $\pm$0.8} & 93.31{\color{grayv} \scriptsize $\pm$2.1} & 92.21{\color{grayv} \scriptsize $\pm$0.5} & 69.50{\color{grayv} \scriptsize $\pm$3.1} & {\textbf{62.05}}{\color{grayv} \scriptsize $\pm$3.0} & 65.26{\color{grayv} \scriptsize $\pm$0.6} & - & .0200 \\
    
    \rowcolor{lightgrayv} \quad + \textbf{\baby} & {\textbf{88.22}}{\color{grayv} \scriptsize $\pm$0.1}$^*$ & {\textbf{79.72}}{\color{grayv} \scriptsize $\pm$0.3}$^*$ & {\textbf{91.04}}{\color{grayv} \scriptsize $\pm$0.2} & {\textbf{94.73}}{\color{grayv} \scriptsize $\pm$0.3}$^*$ & {\textbf{92.85}}{\color{grayv} \scriptsize $\pm$0.0} & {\textbf{73.44}}{\color{grayv} \scriptsize $\pm$0.9}$^*$ & 61.91{\color{grayv} \scriptsize $\pm$1.4} & {\textbf{66.57}}{\color{grayv} \scriptsize $\pm$0.6}$^*$ & {\textbf{1.12}} & {\textbf{.0129}} \\
    
    R\&B + CAFE & 87.16{\color{grayv} \scriptsize $\pm$0.8} & 78.89{\color{grayv} \scriptsize $\pm$0.6} & 90.80{\color{grayv} \scriptsize $\pm$0.7} & 92.73{\color{grayv} \scriptsize $\pm$1.9} & 92.10{\color{grayv} \scriptsize $\pm$0.6} & 68.15{\color{grayv} \scriptsize $\pm$3.4} & {\textbf{62.81}}{\color{grayv} \scriptsize $\pm$3.3} & 65.48{\color{grayv} \scriptsize $\pm$0.9} & - & .0439 \\
    
    \rowcolor{lightgrayv} \quad + \textbf{\baby} & {\textbf{88.38}}{\color{grayv} \scriptsize $\pm$0.3}$^*$ & {\textbf{79.37}}{\color{grayv} \scriptsize $\pm$0.3} & {\textbf{91.02}}{\color{grayv} \scriptsize $\pm$0.6} & {\textbf{94.65}}{\color{grayv} \scriptsize $\pm$1.3}$^*$ & {\textbf{92.78}}{\color{grayv} \scriptsize $\pm$0.2}$^*$ & {\textbf{71.87}}{\color{grayv} \scriptsize $\pm$2.8}$^*$ & 62.75{\color{grayv} \scriptsize $\pm$2.7} & {\textbf{65.99}}{\color{grayv} \scriptsize $\pm$0.8} & {\textbf{1.09}} & {\textbf{.0402}} \\
    
    R\&B + BMR & 87.32{\color{grayv} \scriptsize $\pm$0.3} & 78.87{\color{grayv} \scriptsize $\pm$0.4} & 91.26{\color{grayv} \scriptsize $\pm$0.4} & 93.24{\color{grayv} \scriptsize $\pm$0.8} & 92.23{\color{grayv} \scriptsize $\pm$0.2} & 68.92{\color{grayv} \scriptsize $\pm$1.9} & 62.53{\color{grayv} \scriptsize $\pm$2.5} & 65.51{\color{grayv} \scriptsize $\pm$0.8} & - & .0458 \\
    
    \rowcolor{lightgrayv} \quad + \textbf{\baby} & {\textbf{88.42}}{\color{grayv} \scriptsize $\pm$0.4}$^*$ & {\textbf{79.71}}{\color{grayv} \scriptsize $\pm$0.5}$^*$ & {\textbf{91.42}}{\color{grayv} \scriptsize $\pm$0.8} & {\textbf{94.25}}{\color{grayv} \scriptsize $\pm$1.5}$^*$ & {\textbf{92.90}}{\color{grayv} \scriptsize $\pm$0.3}$^*$ & {\textbf{71.81}}{\color{grayv} \scriptsize $\pm$2.9}$^*$ & {\textbf{62.73}}{\color{grayv} \scriptsize $\pm$1.5} & {\textbf{66.82}}{\color{grayv} \scriptsize $\pm$1.1}$^*$ & {\textbf{1.02}} & {\textbf{.0409}} \\

    R\&B + GAMED & 87.03{\color{grayv} \scriptsize $\pm$0.5} & 78.81{\color{grayv} \scriptsize $\pm$0.3} & {\textbf{91.55}}{\color{grayv} \scriptsize $\pm$0.5} & 92.47{\color{grayv} \scriptsize $\pm$1.4} & 92.01{\color{grayv} \scriptsize $\pm$0.5} & 67.05{\color{grayv} \scriptsize $\pm$0.4} & 64.22{\color{grayv} \scriptsize $\pm$0.3} & 65.60{\color{grayv} \scriptsize $\pm$0.4} & - & .0350 \\
    
    \rowcolor{lightgrayv} \quad + \textbf{\baby} & {\textbf{88.30}}{\color{grayv} \scriptsize $\pm$0.7}$^*$ & {\textbf{79.79}}{\color{grayv} \scriptsize $\pm$0.7}$^*$ & 91.02{\color{grayv} \scriptsize $\pm$0.7} & {\textbf{94.88}}{\color{grayv} \scriptsize $\pm$0.2}$^*$ & {\textbf{92.91}}{\color{grayv} \scriptsize $\pm$0.8}$^*$ & {\textbf{73.88}}{\color{grayv} \scriptsize $\pm$0.5}$^*$ & {\textbf{64.73}}{\color{grayv} \scriptsize $\pm$0.4} & {\textbf{67.67}}{\color{grayv} \scriptsize $\pm$0.7}$^*$ & {\textbf{1.81}} & {\textbf{.0257}} \\
    
    \hline
    \multicolumn{10}{c}{\textbf{Dataset: \textit{Weibo}} \citep{jin2017multimodal} } \\
    
    ResNet + BERT & 90.38{\color{grayv} \scriptsize $\pm$0.7} & 90.37{\color{grayv} \scriptsize $\pm$0.7} & 88.89{\color{grayv} \scriptsize $\pm$1.0} & 91.68{\color{grayv} \scriptsize $\pm$1.2} & 90.22{\color{grayv} \scriptsize $\pm$0.8} & 91.95{\color{grayv} \scriptsize $\pm$1.0} & 89.95{\color{grayv} \scriptsize $\pm$1.1} & 90.53{\color{grayv} \scriptsize $\pm$0.7} & - & .0760 \\
    
    \rowcolor{lightgrayv} \quad + \textbf{\baby} & {\textbf{91.48}}{\color{grayv} \scriptsize $\pm$0.7}$^*$ & {\textbf{91.48}}{\color{grayv} \scriptsize $\pm$0.7}$^*$ & {\textbf{89.98}}{\color{grayv} \scriptsize $\pm$1.4}$^*$ & {\textbf{92.80}}{\color{grayv} \scriptsize $\pm$1.9}$^*$ & {\textbf{91.34}}{\color{grayv} \scriptsize $\pm$0.6}$^*$ & {\textbf{93.09}}{\color{grayv} \scriptsize $\pm$1.6}$^*$ & {\textbf{90.24}}{\color{grayv} \scriptsize $\pm$1.4} & {\textbf{91.61}}{\color{grayv} \scriptsize $\pm$0.8}$^*$ & {\textbf{1.01}} & {\textbf{.0632}} \\

    R\&B + SAFE & 90.52{\color{grayv} \scriptsize $\pm$0.6} & 90.51{\color{grayv} \scriptsize $\pm$0.6} & 89.86{\color{grayv} \scriptsize $\pm$1.2} & 90.68{\color{grayv} \scriptsize $\pm$2.3} & 90.25{\color{grayv} \scriptsize $\pm$0.7} & 91.24{\color{grayv} \scriptsize $\pm$1.9} & 90.36{\color{grayv} \scriptsize $\pm$1.5} & 90.78{\color{grayv} \scriptsize $\pm$0.5} & - & .0629 \\
    
    \rowcolor{lightgrayv} \quad + \textbf{\baby} & {\textbf{91.60}}{\color{grayv} \scriptsize $\pm$0.3}$^*$ & {\textbf{91.59}}{\color{grayv} \scriptsize $\pm$0.3}$^*$ & {\textbf{91.50}}{\color{grayv} \scriptsize $\pm$1.2}$^*$ & {\textbf{91.11}}{\color{grayv} \scriptsize $\pm$1.0} & {\textbf{91.31}}{\color{grayv} \scriptsize $\pm$0.4}$^*$ & {\textbf{91.70}}{\color{grayv} \scriptsize $\pm$1.5} & {\textbf{92.06}}{\color{grayv} \scriptsize $\pm$1.4}$^*$ & {\textbf{91.88}}{\color{grayv} \scriptsize $\pm$0.2}$^*$ & {\textbf{1.07}} & {\textbf{.0583}} \\
    
    R\&B + MCAN & 90.58{\color{grayv} \scriptsize $\pm$0.5} & 90.58{\color{grayv} \scriptsize $\pm$0.5} & 88.43{\color{grayv} \scriptsize $\pm$2.1} & 92.67{\color{grayv} \scriptsize $\pm$1.7} & 90.50{\color{grayv} \scriptsize $\pm$0.4} & 92.80{\color{grayv} \scriptsize $\pm$1.3} & 88.62{\color{grayv} \scriptsize $\pm$2.4} & 90.66{\color{grayv} \scriptsize $\pm$0.6} & - & .0330 \\
    
    \rowcolor{lightgrayv} \quad + \textbf{\baby} & {\textbf{91.76}}{\color{grayv} \scriptsize $\pm$0.3}$^*$ & {\textbf{91.76}}{\color{grayv} \scriptsize $\pm$0.3}$^*$ & {\textbf{90.48}}{\color{grayv} \scriptsize $\pm$1.0}$^*$ & {\textbf{92.77}}{\color{grayv} \scriptsize $\pm$1.8} & {\textbf{91.59}}{\color{grayv} \scriptsize $\pm$0.4}$^*$ & {\textbf{93.10}}{\color{grayv} \scriptsize $\pm$1.5} & {\textbf{90.81}}{\color{grayv} \scriptsize $\pm$1.3}$^*$ & {\textbf{91.92}}{\color{grayv} \scriptsize $\pm$0.2}$^*$ & {\textbf{1.17}} & {\textbf{.0280}} \\
    
    R\&B + CAFE & 90.72{\color{grayv} \scriptsize $\pm$0.5} & 90.71{\color{grayv} \scriptsize $\pm$0.4} & 89.19{\color{grayv} \scriptsize $\pm$1.5} & 91.96{\color{grayv} \scriptsize $\pm$1.1} & 90.56{\color{grayv} \scriptsize $\pm$0.4} & 92.23{\color{grayv} \scriptsize $\pm$0.9} & 89.55{\color{grayv} \scriptsize $\pm$1.8} & 90.87{\color{grayv} \scriptsize $\pm$0.5} & - & .0952 \\
    
    \rowcolor{lightgrayv} \quad + \textbf{\baby} & {\textbf{91.76}}{\color{grayv} \scriptsize $\pm$0.2}$^*$ & {\textbf{91.76}}{\color{grayv} \scriptsize $\pm$0.2}$^*$ & {\textbf{89.53}}{\color{grayv} \scriptsize $\pm$1.4} & {\textbf{94.02}}{\color{grayv} \scriptsize $\pm$1.4}$^*$ & {\textbf{91.70}}{\color{grayv} \scriptsize $\pm$0.2}$^*$ & {\textbf{94.14}}{\color{grayv} \scriptsize $\pm$1.2}$^*$ & {\textbf{89.65}}{\color{grayv} \scriptsize $\pm$1.6} & {\textbf{91.82}}{\color{grayv} \scriptsize $\pm$0.3}$^*$ & {\textbf{1.07}} & {\textbf{.0773}} \\
    
    R\&B + BMR & 90.86{\color{grayv} \scriptsize $\pm$0.7} & 90.86{\color{grayv} \scriptsize $\pm$0.7} & 89.11{\color{grayv} \scriptsize $\pm$2.2} & 92.52{\color{grayv} \scriptsize $\pm$1.8} & 90.75{\color{grayv} \scriptsize $\pm$0.6} & 92.77{\color{grayv} \scriptsize $\pm$1.4} & 89.31{\color{grayv} \scriptsize $\pm$2.7} & 90.97{\color{grayv} \scriptsize $\pm$0.9} & - & .0119 \\
    
    \rowcolor{lightgrayv} \quad + \textbf{\baby} & {\textbf{92.01}}{\color{grayv} \scriptsize $\pm$1.6}$^*$ & {\textbf{92.01}}{\color{grayv} \scriptsize $\pm$1.6}$^*$ & {\textbf{90.44}}{\color{grayv} \scriptsize $\pm$0.8}$^*$ & {\textbf{93.37}}{\color{grayv} \scriptsize $\pm$2.7}$^*$ & {\textbf{91.88}}{\color{grayv} \scriptsize $\pm$0.9}$^*$ & {\textbf{93.59}}{\color{grayv} \scriptsize $\pm$2.7}$^*$ & {\textbf{90.74}}{\color{grayv} \scriptsize $\pm$2.2}$^*$ & {\textbf{92.14}}{\color{grayv} \scriptsize $\pm$1.3}$^*$ & {\textbf{1.13}} & {\textbf{.0101}} \\

    R\&B + GAMED & 90.24{\color{grayv} \scriptsize $\pm$0.5} & 90.19{\color{grayv} \scriptsize $\pm$0.5} & 89.27{\color{grayv} \scriptsize $\pm$1.4} & 91.04{\color{grayv} \scriptsize $\pm$1.6} & 89.51{\color{grayv} \scriptsize $\pm$0.7} & 91.79{\color{grayv} \scriptsize $\pm$1.8} & 89.18{\color{grayv} \scriptsize $\pm$1.8} & 90.87{\color{grayv} \scriptsize $\pm$0.3} & - & .0798 \\
    
    \rowcolor{lightgrayv} \quad + \textbf{\baby} & {\textbf{92.08}}{\color{grayv} \scriptsize $\pm$0.2}$^*$ & {\textbf{92.08}}{\color{grayv} \scriptsize $\pm$0.2}$^*$ & {\textbf{90.34}}{\color{grayv} \scriptsize $\pm$1.4}$^*$ & {\textbf{93.65}}{\color{grayv} \scriptsize $\pm$1.4}$^*$ & {\textbf{91.97}}{\color{grayv} \scriptsize $\pm$0.2}$^*$ & {\textbf{93.84}}{\color{grayv} \scriptsize $\pm$1.2}$^*$ & {\textbf{90.61}}{\color{grayv} \scriptsize $\pm$1.6}$^*$ & {\textbf{92.19}}{\color{grayv} \scriptsize $\pm$0.3}$^*$ & {\textbf{1.84}} & {\textbf{.0616}} \\

    \hline
    \multicolumn{10}{c}{\textbf{Dataset: \textit{Twitter}} \citep{boididou2018detection} } \\
    
    ResNet + BERT & 66.02{\color{grayv} \scriptsize $\pm$1.9} & 65.84{\color{grayv} \scriptsize $\pm$1.8} & 58.40{\color{grayv} \scriptsize $\pm$2.4} & {\textbf{68.09}}{\color{grayv} \scriptsize $\pm$3.9} & {\textbf{63.63}}{\color{grayv} \scriptsize $\pm$1.6} & 74.14{\color{grayv} \scriptsize $\pm$1.7} & 63.02{\color{grayv} \scriptsize $\pm$3.9} & 68.04{\color{grayv} \scriptsize $\pm$2.7} & - & .1428 \\
    
    \rowcolor{lightgrayv} \quad + \textbf{\baby} & {\textbf{68.15}}{\color{grayv} \scriptsize $\pm$2.4}$^*$ & {\textbf{67.46}}{\color{grayv} \scriptsize $\pm$3.9}$^*$ & {\textbf{61.46}}{\color{grayv} \scriptsize $\pm$3.5}$^*$ & 66.45{\color{grayv} \scriptsize $\pm$4.4} & 63.49{\color{grayv} \scriptsize $\pm$3.7} & {\textbf{74.40}}{\color{grayv} \scriptsize $\pm$2.3} & {\textbf{69.40}}{\color{grayv} \scriptsize $\pm$2.8}$^*$ & {\textbf{71.46}}{\color{grayv} \scriptsize $\pm$2.3}$^*$ & {\textbf{1.89}} & {\textbf{.1002}} \\

    R\&B + SAFE & 66.41{\color{grayv} \scriptsize $\pm$3.5} & 66.13{\color{grayv} \scriptsize $\pm$3.5} & 58.69{\color{grayv} \scriptsize $\pm$3.6} & 66.61{\color{grayv} \scriptsize $\pm$3.7} & 62.43{\color{grayv} \scriptsize $\pm$3.7} & 72.62{\color{grayv} \scriptsize $\pm$3.2} & 63.01{\color{grayv} \scriptsize $\pm$3.8} & 68.24{\color{grayv} \scriptsize $\pm$3.6} & - & .0927 \\
    
    \rowcolor{lightgrayv} \quad + \textbf{\baby} & {\textbf{68.28}}{\color{grayv} \scriptsize $\pm$2.6}$^*$ & {\textbf{68.03}}{\color{grayv} \scriptsize $\pm$2.5}$^*$ & {\textbf{60.87}}{\color{grayv} \scriptsize $\pm$2.6}$^*$ & {\textbf{72.32}}{\color{grayv} \scriptsize $\pm$4.1}$^*$ & {\textbf{65.86}}{\color{grayv} \scriptsize $\pm$2.7}$^*$ & {\textbf{76.43}}{\color{grayv} \scriptsize $\pm$3.5}$^*$ & {\textbf{65.31}}{\color{grayv} \scriptsize $\pm$3.3}$^*$ & {\textbf{70.19}}{\color{grayv} \scriptsize $\pm$3.1}$^*$ & {\textbf{2.89}} & {\textbf{.0707}} \\

    R\&B + MCAN & 66.66{\color{grayv} \scriptsize $\pm$2.4} & 66.50{\color{grayv} \scriptsize $\pm$2.3} & 59.02{\color{grayv} \scriptsize $\pm$2.9} & 68.88{\color{grayv} \scriptsize $\pm$1.6} & 64.37{\color{grayv} \scriptsize $\pm$1.6} & 74.72{\color{grayv} \scriptsize $\pm$1.3} & 63.56{\color{grayv} \scriptsize $\pm$2.7} & 68.64{\color{grayv} \scriptsize $\pm$3.1} & - & .0459 \\
    
    \rowcolor{lightgrayv} \quad + \textbf{\baby} & {\textbf{68.60}}{\color{grayv} \scriptsize $\pm$2.1}$^*$ & {\textbf{68.43}}{\color{grayv} \scriptsize $\pm$1.8}$^*$ & {\textbf{61.02}}{\color{grayv} \scriptsize $\pm$3.0}$^*$ & {\textbf{75.26}}{\color{grayv} \scriptsize $\pm$2.9}$^*$ & {\textbf{67.12}}{\color{grayv} \scriptsize $\pm$1.5}$^*$ & {\textbf{77.79}}{\color{grayv} \scriptsize $\pm$1.4}$^*$ & {\textbf{63.68}}{\color{grayv} \scriptsize $\pm$3.5} & {\textbf{69.75}}{\color{grayv} \scriptsize $\pm$2.4}$^*$ & {\textbf{2.41}} & {\textbf{.0382}} \\
    
    R\&B + CAFE & 66.73{\color{grayv} \scriptsize $\pm$1.2} & 66.35{\color{grayv} \scriptsize $\pm$1.1} & 59.74{\color{grayv} \scriptsize $\pm$1.7} & 66.75{\color{grayv} \scriptsize $\pm$4.1} & 62.97{\color{grayv} \scriptsize $\pm$1.5} & 73.20{\color{grayv} \scriptsize $\pm$1.4} & 66.71{\color{grayv} \scriptsize $\pm$4.1} & 69.73{\color{grayv} \scriptsize $\pm$1.1} & - & .0872 \\
    
    \rowcolor{lightgrayv} \quad + \textbf{\baby} & {\textbf{69.16}}{\color{grayv} \scriptsize $\pm$1.7}$^*$ & {\textbf{68.73}}{\color{grayv} \scriptsize $\pm$1.7}$^*$ & {\textbf{62.66}}{\color{grayv} \scriptsize $\pm$2.4}$^*$ & {\textbf{68.52}}{\color{grayv} \scriptsize $\pm$3.2}$^*$ & {\textbf{65.31}}{\color{grayv} \scriptsize $\pm$1.8}$^*$ & {\textbf{75.11}}{\color{grayv} \scriptsize $\pm$1.4}$^*$ & {\textbf{69.64}}{\color{grayv} \scriptsize $\pm$2.9}$^*$ & {\textbf{72.15}}{\color{grayv} \scriptsize $\pm$1.7}$^*$ & {\textbf{2.39}} & {\textbf{.0663}} \\
    
    R\&B + BMR & 66.71{\color{grayv} \scriptsize $\pm$1.4} & 66.33{\color{grayv} \scriptsize $\pm$1.5} & 59.62{\color{grayv} \scriptsize $\pm$1.4} & 66.79{\color{grayv} \scriptsize $\pm$3.8} & 62.94{\color{grayv} \scriptsize $\pm$2.5} & 73.24{\color{grayv} \scriptsize $\pm$2.2} & 66.65{\color{grayv} \scriptsize $\pm$2.6} & 69.73{\color{grayv} \scriptsize $\pm$1.2} & - & .0145 \\
    
    \rowcolor{lightgrayv} \quad + \textbf{\baby} & {\textbf{69.57}}{\color{grayv} \scriptsize $\pm$2.7}$^*$ & {\textbf{69.00}}{\color{grayv} \scriptsize $\pm$2.3}$^*$ & {\textbf{64.29}}{\color{grayv} \scriptsize $\pm$3.5}$^*$ & {\textbf{68.39}}{\color{grayv} \scriptsize $\pm$3.6}$^*$ & {\textbf{65.71}}{\color{grayv} \scriptsize $\pm$2.2}$^*$ & {\textbf{75.24}}{\color{grayv} \scriptsize $\pm$2.1}$^*$ & {\textbf{70.45}}{\color{grayv} \scriptsize $\pm$4.3}$^*$ & {\textbf{72.30}}{\color{grayv} \scriptsize $\pm$3.1}$^*$ & {\textbf{2.87}} & {\textbf{.0137}} \\

    R\&B + GAMED & 66.53{\color{grayv} \scriptsize $\pm$2.5} & 66.21{\color{grayv} \scriptsize $\pm$2.5} & 58.93{\color{grayv} \scriptsize $\pm$2.0} & 66.61{\color{grayv} \scriptsize $\pm$2.5} & 62.57{\color{grayv} \scriptsize $\pm$2.4} & 73.25{\color{grayv} \scriptsize $\pm$2.2} & 63.52{\color{grayv} \scriptsize $\pm$2.3} & 68.84{\color{grayv} \scriptsize $\pm$2.6} & - & .1122 \\
    
    \rowcolor{lightgrayv} \quad + \textbf{\baby} & {\textbf{69.77}}{\color{grayv} \scriptsize $\pm$2.5}$^*$ & {\textbf{69.77}}{\color{grayv} \scriptsize $\pm$2.3}$^*$ & {\textbf{63.96}}{\color{grayv} \scriptsize $\pm$1.9}$^*$ & {\textbf{69.11}}{\color{grayv} \scriptsize $\pm$1.2}$^*$ & {\textbf{65.57}}{\color{grayv} \scriptsize $\pm$2.4}$^*$ & {\textbf{75.42}}{\color{grayv} \scriptsize $\pm$2.5}$^*$ & {\textbf{69.41}}{\color{grayv} \scriptsize $\pm$2.0}$^*$ & {\textbf{71.97}}{\color{grayv} \scriptsize $\pm$1.8}$^*$ & {\textbf{3.56}} & {\textbf{.0838}} \\
    
    \bottomrule
  \end{tabular} }
\end{table*}

\vspace{2pt} \noindent
\textbf{Dependency relationship.}
The augmented images correspond to text segments that not only exhibit inherent temporal relationships, but also logical dependency relationships, \eg coordination relationships. To integrate these dependencies into the graph $\mathfrak{G}$, we first utilize the prevalent spaCy toolkit to construct a dependency tree structure from the original text. These dependency tree structures typically represent token-level relationships, therefore, to consolidate these into text segment-level relationships, we merge all token nodes within a segment into a single node while retaining the connected edges. Based on this operation, we can obtain a new segment-level graph structure and incorporate it into $\mathfrak{G}$.

Upon these relationships, we obtain a graph structure $\mathfrak{G}$, and we use a graph neural network to encode a fused image feature as follows:
\begin{equation}
    \label{eq11}
    \mathbf{e}_i^{v(l)} = \boldsymbol{\sigma} \left( \sum \nolimits _{j=1}^{K + 1} \boldsymbol{\Lambda}_{ij} \boldsymbol{\theta}^{g(l)} \mathbf{e}_i^{v(l - 1)} + b^{(l)} \right),
\end{equation}
where $\boldsymbol{\sigma}(\cdot)$ is a activation function, and $\boldsymbol{\theta}^{g(l)}$ and $b^{(l)}$ are a linear weight matrix and a bias term, respectively. And we initialize $\mathbf{e}_i^{v(0)}$ as $\mathbf{H}_i$, and denote the feature $\mathbf{e}_i^{v(l)}$ at the final GCN layer as $\mathbf{e}_i^{v}$. In our model, we empirically fix the number of GCN layers to 2.
Given the fused feature $\mathbf{e}_i^{v}$ and text and image features $\mathbf{h}_i^t$ and $\mathbf{h}_i^v$, we employ a cross-attention network to obtain the multimodal feature as follows:
\begin{equation}
    \label{eq12}
    \begin{aligned}
        \mathbf{e}_i & = \boldsymbol{\mu} \left( \mathbf{o}_i \mathbf{W}^Q \big( \mathbf{h}_i^t \mathbf{W}^K \big)^\top / \sqrt{d_n} \right) \mathbf{o}_i \mathbf{W}^V, \\ 
        \mathbf{o}_i & = \boldsymbol{\mu} \left( \mathbf{e}_i^v \mathbf{W}^Q \big( \mathbf{h}_i^v \mathbf{W}^K \big)^\top / \sqrt{d_n} \right) \mathbf{e}_i^v \mathbf{W}^V,
    \end{aligned}
\end{equation}
where $\boldsymbol{\theta}^c = \{\mathbf{W}^Q, \mathbf{W}^K, \mathbf{W}^V\}$ represents the parameter of the attention network.

%% file: S_Experiment.tex
\section{Experiments} \label{sec:experiment}

In this section, we evaluate the performance of our proposed \baby by answering the following questions:
\begin{itemize}
    \item \textbf{EQ1.} How does \baby perform on MMD?
    \item \textbf{EQ2.} Are all modules in \baby effective?
    \item \textbf{EQ3.} Can \baby enhance the contribution of the image modality to MMD?
\end{itemize}
For clarity, the experimental settings about datasets, baselines and experimental details can be seen in the Appendix.

\subsection{Main Results (EQ1, EQ3)}

We select five baseline models to compare the performance of \baby across three prevalent MMD datasets. Meanwhile, we employ nine metrics to evaluate the prediction capability of the model, where eight metrics, \eg accuracy (Acc.) and F1 score, assess the accuracy between the model’s predictions and the gold veracity labels, and one metric $\mathscr{G}(y, \mathbf{x}^v)$ represents the information gain brought by $\mathbf{x}^v$ to quantitatively measure our model’s ability to improve the image modality contribution. The experimental results are reported in Table~\ref{result}.
Generally, \baby surpasses its baseline models over most datasets and metrics. For example, on the \textit{Twitter} dataset, \baby outperforms its baseline model BMR, by approximately 2.86 on the average of all metrics. This sufficiently demonstrates the effectiveness of the method proposed in this paper for enhancing the detection performance of MMD models.

Turning to the quantitative metric $\mathscr{G}(y, \mathbf{x}^v)$, our method outperforms the baseline models on most datasets. This result demonstrates that our method can improve the image modality contribution to MMD and enable the model to give more confident predictions by introducing additional augmented images. Exceptionally, $\mathscr{G}(y, \mathbf{x}^v)$ did not achieve the optimal result on the BMR model across \textit{Twitter}. Upon observing the results of the experiments, we find that this model, when the text modality is removed, tends to predict all samples as the real class with high confidence, resulting in a low prediction entropy. However, our method mitigates this tendency, further validating its effectiveness.

\begin{table}[t]
\centering
\renewcommand\arraystretch{0.90}
  \caption{Ablation study on three ablative versions.}
  \label{ablation}
  \footnotesize
  \setlength{\tabcolsep}{5pt}{
  \begin{tabular}{m{1.5cm}m{0.8cm}<{\centering}m{0.8cm}<{\centering}m{0.8cm}<{\centering}m{0.8cm}<{\centering}m{1.0cm}<{\centering}}
    \toprule
    \quad Model & Acc. & F1 & F1$_{\text{real}}$ & F1$_{\text{fake}}$ & Avg.$\boldsymbol{\Delta}$ \\
    \hline
    \multicolumn{6}{c}{\textbf{Dataset}: \textit{GossipCop}} \\
    \rowcolor{lightgrayv} \textbf{\baby} & 88.42 & 79.71 & 92.90 & 66.82 & / \\
    \quad \textit{w/o} GF & 87.67 & 79.06 & 92.34 & 66.27 & \textbf{0.63} \\
    \quad \textit{w/o} MI & 87.60 & 78.93 & 92.35 & 65.82 & \textbf{0.79} \\
    \quad \textit{w/o} AI & 87.32 & 78.87 & 92.23 & 65.51 & \textbf{0.98} \\
    \hline
    \multicolumn{6}{c}{\textbf{Dataset}: \textit{Weibo}} \\
    \rowcolor{lightgrayv} \textbf{\baby} & 92.01 & 92.01 & 91.88 & 92.14 & / \\
    \quad \textit{w/o} GF & 91.33 & 91.33 & 91.17 & 91.48 & \textbf{0.68} \\
    \quad \textit{w/o} MI & 91.06 & 91.05 & 90.92 & 91.38 & \textbf{0.91} \\
    \quad \textit{w/o} AI & 90.86 & 90.86 & 90.75 & 90.97 & \textbf{1.15} \\
    \hline
    \multicolumn{6}{c}{\textbf{Dataset}: \textit{Twitter}} \\
    \rowcolor{lightgrayv} \textbf{\baby} & 69.57 & 69.00 & 65.71 & 72.30 & / \\
    \quad \textit{w/o} GF & 67.86 & 67.41 & 63.55 & 71.26 & \textbf{1.63} \\
    \quad \textit{w/o} MI & 67.18 & 66.94 & 63.33 & 70.54 & \textbf{2.15} \\
    \quad \textit{w/o} AI & 66.71 & 66.33 & 62.94 & 69.73 & \textbf{2.72} \\
    \bottomrule
  \end{tabular} }
\end{table}

\subsection{Ablative Study (EQ2)}

To answer EQ2, we conduct the ablative study on three datasets, and the results are shown in Table~\ref{ablation}. Specifically, we investigate the following four ablative versions:
\begin{itemize}
    \item \textbf{w/o graph fusion (GF)} $\mathfrak{G}$ directly concatenate three features $\mathbf{h}^t$, $\mathbf{h}^v$, and $\mathbf{h}^g$, instead of fusing them with a graph $\mathfrak{G}$ and the graph neural network;
    \item \textbf{w/o mutual information (MI)} $\mathcal{R}_{MTI}$ \textbf{and} $\mathcal{R}_{MIL}$ does not utilize mutual information regularizations $\mathcal{R}_{MTI}$ and $\mathcal{R}_{MIL}$ to tune the text-to-image generator;
    \item \textbf{w/o augmented image (AI)} $\mathbf{x}^g$ removes the augmented images to enhance the MMD model, which is equivalent to the baseline model compared in Table~\ref{result}.
\end{itemize}

In Table~\ref{ablation}, we report the decrease in accuracy for four variants compared to the BMR baseline model. In general, the model's performance consistently declines when any module is removed, demonstrating the importance of these modules. Meanwhile, in more detail, the accuracy ranking of the four variants is consistently w/o graph fusion \textgreater w/o mutual information \textgreater w/o augmented image.
Specifically, 
removing the mutual information regularizations, although still outperforming the baseline model, shows a significant decrease in performance compared to the full version of \baby. This suggests that our designed mutual information method can indeed significantly increase the information \wrt the veracity label in the generated images, thereby improving detection performance.
The graph fusion utilizes the three types of relationships to construct and encode the graph, and the ablative results also demonstrate the effectiveness of this fusion approach.

%% file: S_Relatedworks.tex
\section{Related Works} \label{sec:related}

Generally, recent MMD methods achieve this by learning the potential relationships between the multimodal content and their veracity labels \citep{wu2023see,hu2023causal,ying2023bootstrapping,wang2024harmfully}. Specifically, these methods present various external features \citep{wang2024harmfully,dong2024unveiling}, inconsistency measures \citep{chen2022cross,ma2024event}, knowledge-augmented methods \citep{fung2021infosurgeon,xuan2024lemma}, and multimodal fusion strategies \citep{ying2023bootstrapping,wang2024escaping,wang2024fake} to enhance the models. For example, \citet{wang2024harmfully} capture image manipulation features and their intention features to identify harmfully manipulated images in MMD datasets; \citet{ma2024event} learn the event-level inconsistency by constructing a dependency graph structure and utilize its feature with a multi-view learning framework. Additionally, with the development of large vision-language models, recent MMD works use these models to enhance detection models by synthesizing data \citep{zeng2024multimodal}, integrating evidence \citep{tahmasebi2024multimodal} and supplementing knowledge \citep{liu2024fka,xuan2024lemma}.

Unlike these MMD methods, we observe a problem that the image modality contributes less in existing MMD models . In the community, a few MMD efforts focus on similar phenomena. For example, \citet{chen2023causal} found that only using images to detect fake news is unreliable, hence they proposed a casual approach. Nevertheless, a comprehensive empirical study on this issue remains a blank.

%% file: S_Conclusion.tex
\section{Conclusion}

In this work, we empirically observe that the text modality contributes more to MMD than the image modality since the text describes the whole story of the MMD post but the image always presents partial scenes only. Therefore, to boost the contribution of the image modality, we propose a new MMD framework \baby, which aims to generate a sequence of augmented images that replay the whole story in the text. To achieve this goal, we split the text into a sequence of segments, which present partial scenes, and feed them into a pre-trained text-to-image generator to generate a sequence of augmented images.
To ensure the quality of the augmented images, we tune the generator with two text-image and image-label mutual information. Meanwhile, to effectively integrate the image features, we construct a graph with their potential relationships and employ a graph neural network to fuse them. The experiments can demonstrate that \baby improves the performance of the baseline models, and alleviates the information gap between text and images.

%% file: S_Appendix.tex
\appendix

\section{Overall Algorithm of \baby} \label{sec:appendix.algorithm}

The complete algorithm of \baby is presented in Alg.~\ref{algorithm}. Due to the cost of training diffusion models and using the model for inference, we have additionally use two new hyper-parameters: update step $T_u$ and generation step $T_g$. While training the detector $\mathcal{F}_{\boldsymbol{\theta}}(\cdot)$, if the training reaches an iteration number that is a multiple of the update step, the model will perform an update of the generator $\mathcal{G}_{\boldsymbol{\phi}}(\cdot)$ using Eq.(4); if it reaches an iteration number that is a multiple of the generation step, we will re-generate images for the text in the entire MMD dataset $\mathcal{D}$ and replace the images generated in the previous generation step.

\renewcommand{\algorithmicrequire}{\textbf{Input:}}
\renewcommand{\algorithmicensure}{\textbf{Output:}}
\begin{algorithm}[h]
    \caption{Training pipeline of \baby.}
    \label{algorithm}
    \begin{algorithmic}[1]
    \Require MMD dataset $\mathcal{D}$; 
    text-to-image generation dataset $\overline{\mathcal{D}}$; 
    initialized detector $\boldsymbol{\theta}$ by the weights of BERT and ResNet, and random weights;
    initialized generator $\boldsymbol{\phi}$ by the weights of stable diffusion;
    hyper-paramaters $\alpha_1$, $\alpha_2$, and $\beta$; 
    update step $T_u$, generation step $T_g$;
    number pf training iterations $I$.
    \Ensure Tuned MMD model parameterized by $\boldsymbol{\theta}$.
    \State Generate initial images $\mathbf{x}^g$ with $\boldsymbol{\phi}$;
    \For{$i \in \{1, 2, \cdots, I\}$}
    \State Draw mini-batch $\mathcal{B}$ from $\mathcal{D}$ randomly;
    \State Optimize $\boldsymbol{\theta}$ with $\mathcal{L}_{DET}$ in Eq.(6);
        \If{$i \ \% \  T_g = 0$}
        \State Generate images with $\boldsymbol{\phi}$ and update previous $\mathbf{x}^g$;
        \EndIf
        \If{$i \ \% \  T_u = 0$}
        \State Draw mini-batches $\overline{\mathcal{B}}$ from $\overline{\mathcal{D}}$ randomly;
        \State Optimize $\boldsymbol{\phi}$ with $\mathcal{L}_{GEN}$ in Eq.(4).
        \EndIf
    \EndFor
    \end{algorithmic}
\end{algorithm}

\section{Experimental Settings} \label{sec:appendix.settings}

In this section, we describe the experimental settings about our used benchmark MMD datasets, SOTA MMD models, and implementation details.

\subsection{Benchmark Datasets} \label{sec:appendix.settings.datasets}

We use the following three MMD datasets to evaluate \baby. For clarity, the statistics of these datasets are depicted in Table~\ref{datasetsta}.

\begin{itemize}
    \item \textbf{\textit{GossipCop}} \citep{shu2020fakenewsnet} is a subset of the comprehensive \textit{FakeNewsNet} repository, which contains 12,840 text-image pairs. Its topics mainly focus on celebrity gossip and entertainment news, so the majority of the images contained therein are photographs of individuals.
    \item \textbf{\textit{Weibo}} \citep{jin2017multimodal} is a Chinese MMD dataset comprising 9,528 text-image pairs. Its real-class samples originate from an authoritative news media source \textit{Xinhua.net}, while the fake-class samples are sourced from a misinformation disclosing platform developed by \textit{Weibo.com}. The majority of the samples in the dataset pertain to social hotspots and are derived from social media posts.
    \item \textbf{\textit{Twitter}} \citep{boididou2018detection} is a dataset collected from the social media platform, characterized by multiple texts corresponding to a single image. Specifically, it includes 13,924 text entries but only 514 images.
\end{itemize}

\begin{table}[t]
\centering
\renewcommand\arraystretch{1.0}
  \caption{Statistics of MMD datasets. \#\textit{R.}, \#\textit{F.}, \#\textit{Img.}, and \#\textit{Avg.L} represent the number of real-class samples, fake-class samples, images, and average article length, respectively.}
  \label{datasetsta}
  \setlength{\tabcolsep}{5pt}{
  \begin{tabular}{m{2.0cm}<{\centering}m{1.0cm}<{\centering}m{1.0cm}<{\centering}m{1.0cm}<{\centering}m{1.0cm}<{\centering}}
    \toprule
    Dataset & \#\textit{R.} & \#\textit{F.} & \#\textit{Img.} & \#\textit{Avg.L} \\
    \hline
    \textit{GossipCop} & 10,259 & 2,581 & 12,840 & 566.2 \\
    \textit{Weibo} & 4,779 & 4,749 & 9,528 & 102.9 \\
    \textit{Twitter} & 6,026 & 7,898 & 514 & 16.5 \\
    \bottomrule
  \end{tabular} }
\end{table}

\subsection{MMD Models} \label{sec:appendix.settings.models}

The descriptions of the baseline models used in our empirical observation (Sec.2) and evaluation (Sec.4) sections are as follows:

\begin{itemize}
    \item \textbf{R\&B and M\&D models}. These models first employs pre-trained language models, \eg BERT \citep{devlin2019bert}, and visual models, \eg ResNet \citep{he2016deep}, to encode text and image data, respectively. Then, a linear layer or specific regularization techniques are used to align and fuse the features from different modalities, \eg direct concatenation. The fused features are then fed into a veracity classifier for prediction. In our experiments, we utilize ResNet\footnote{\url{https://download.pytorch.org/models/resnet34-333f7ec4.pth}} + BERT\footnote{\url{https://huggingface.co/google-bert/bert-base-uncased}} (R\&B) \citep{he2016deep,devlin2019bert}, and MAE\footnote{\url{https://huggingface.co/facebook/vit-mae-base}} + DeBERTaV3\footnote{\url{https://huggingface.co/microsoft/deberta-v3-base}} (M\&D) \citep{he2022masked,he2023debertav3} as representatives of the dual-tower model. BERT, DeBERTaV3, and MAE all employ the Transformer architecture \citep{vaswani2017attention} and are pre-trained using masked token prediction or replaced token detection strategies. 
    \item \textbf{CLIP}\footnote{\url{https://huggingface.co/openai/clip-vit-large-patch14}} \citep{radford2021learning} is pre-trained with a contrastive learning objective. In contrast to the above approach, the CLIP model employs a single structure to complete the processes of feature extraction, alignment, and fusion. It segments text by word tokens and images by patches, concatenates and inputs them into a Transformer model, directly obtaining a fused feature.
    \item \textbf{SAFE} \citep{zhou2020safe} is a prevalent MMD model that builds upon the R\&B model by incorporating a cross-modal similarity as a new feature.
    \item \textbf{MCAN} \citep{wu2021multimodal} focuses on improving the multimodal feature fusion module by converting the self-attention mechanism in the Transformer model to an inter-attention mechanism among image, frequency-domain image, and text for feature fusion.
    \item \textbf{CAFE} \citep{chen2022cross} focuses on learning the inconsistencies between modalities, arguing that the importance of unimodal features and multimodal features for veracity decision varies across different samples. Therefore, it designs an inconsistency metric based on a variational encoder and uses this metric as the weight in feature fusion.
    \item \textbf{BMR} \citep{ying2023bootstrapping} is an MMD model, which uses a mixture-of-expert model to boost the model’s feature extraction capability and employs them to construct multi-view news representations.
    \item \textbf{GAMED} \citep{shen2025gamed} is the latest MMD model, which modifies the multi-view feature extraction component of the BMR model to a more effective variational encoder. Additionally, it introduces an AdaIN strategy to sample features from the variational distribution.
\end{itemize}

\subsection{Implementation details} \label{sec:appendix.settings.details}

In our data processing phase, we resize all images, including those from the MMD datasets and the text-to-image generation dataset, to a resolution of 256 $\times$ 256, and randomly crop regions of 224 $\times$ 224. For all text data, we truncated the text from the MMD dataset to a length of 128 tokens, and the text from the text-to-image generation dataset to 77 tokens.
For our models, we use the pre-trained stable diffusion model\footnote{\url{https://huggingface.co/CompVis/stable-diffusion-v1-4}} \citep{rombach2022high} with low-rank adaptation models (rank = 4) \citep{hu2022lora} as our text-to-image generator, and we tune it with the \textit{LAION-2B} dataset \citep{schuhmann2022laion} filtered by aesthetic scores.\footnote{\url{https://huggingface.co/datasets/liangyuch/laion2B-en-aesthetic-seed}}

During the training phase, we utilize the Adam optimizer with a learning rate of $3 \times 10^{-5}$ to train BERT-series models, the AdamW optimizer with a learning rate of $1 \times 10^{-4}$ to train the text-to-image generator, and for other modules, we employ the Adam optimizer with a learning rate of $1 \times 10^{-3}$. We set the batch size to 16 for training the misinformation detector and to 4 for training the text-to-image generator. Throughout the training of the misinformation detector, we tune the text-to-image generator once every 5 epochs and use it to update the generated images. Additionally, we employ an early stopping strategy to constrain the number of training epochs for the detector, which terminates training when the model fails to achieve a better Micro F1 score for continuously 10 epochs on the validation set. Regarding the hyperparameters during training, we empirically fix $\alpha_1$, $\alpha_2$, and $\beta$ to 0.01, and $K$ to 5. To avoid the randomness, we repeat the experiments 5 times with 5 different seeds $\{1, 2, 3, 4, 5\}$, and report their average scores.

\section{More Experimental Results}

In this section, we conduct two additional experiments to validate the quality of the generated images and perform a sensitivity analysis about the number of text segments.

\subsection{Image Quality Evaluation}

In addition to using detection accuracy and modality contribution degree to quantitatively evaluate the quality of generated images, we also attempt to evaluate the generated images using several quantitative metrics commonly employed in the community of image quality evaluation. Specifically, we utilize three metrics, including two no-reference metrics \textbf{NIQE} \citep{mittal2013making} and \textbf{PIQE} \citep{n2015blind}, and one similarity metric \textbf{SIM}, which computes the average similarity between the original text segmentation and the generated images via CLIP \citep{radford2021learning}. We compare the original images, the outputs from a diffusion model without fine-tuning, and our fine-tuned text-to-image generation model. The experimental results are listed in Table~\ref{quality}.

\begin{table}[t]
\centering
\renewcommand\arraystretch{1.10}
  \caption{Evaluation of the image quality. Origin and SD represent the original images in the dataset and images generated by the stable diffusion model without fine-tuning.}
  \label{quality}
  \setlength{\tabcolsep}{5pt}{
  \begin{tabular}{m{1.7cm}<{\centering}m{1.3cm}<{\centering}m{1.0cm}<{\centering}m{1.0cm}<{\centering}m{1.4cm}<{\centering}}
    \toprule
    Dataset & Metric & Origin & SD & \baby \\
    \hline
    \multirow{3}{*}{\textit{GossipCop}}
    & NIQE & 3.73 & 5.28 & \textbf{3.68} \\
    & PIQE & 31.17 & 35.88 & \textbf{30.30} \\
    & SIM & - & 0.427 & \textbf{0.540} \\
    \hline
    \multirow{3}{*}{\textit{Weibo}}
    & NIQE & \textbf{4.40} & 6.01 & 4.52 \\
    & PIQE & \textbf{41.92} & 45.95 & 42.35 \\
    & SIM & - & 0.525 & \textbf{0.573} \\
    \hline
    \multirow{3}{*}{\textit{Twitter}}
    & NIQE & 4.02 & 5.65 & \textbf{3.87} \\
    & PIQE & \textbf{33.50} & 38.71 & 33.72 \\
    & SIM & - & 0.482 & \textbf{0.507} \\
    \bottomrule
  \end{tabular} }
\end{table}

Generally, the quality of the images generated by our method \baby, is comparable to that of the original images in the dataset, and even surpasses them in certain metrics. This result is attributable to our continual training of the generator. However, when compared with the diffusion model without fine-tuning, the image quality produced by our method consistently and significantly outperforms it, further demonstrating the effectiveness of our designed training objectives. The similarity metric SIM quantifies the semantic similarity between the generated images and their corresponding text segments. The results also show that our method significantly surpasses the performance of the generator without fine-tuning. 
Turning to compare metrics across different datasets, we observe that the ranking of image quality among the three datasets is approximately \textit{GossipCop} $>$ \textit{Twitter} $>$ \textit{Weibo}. This is because the \textit{GossipCop} dataset primarily consists of carefully curated images from the entertainment domain, which are generally of higher quality. In contrast, the \textit{Weibo} and \textit{Twitter} datasets contain a large number of manipulated or meaningless images, resulting in lower overall quality.

\begin{figure}[t]
  \centering
  \includegraphics[scale=0.18]{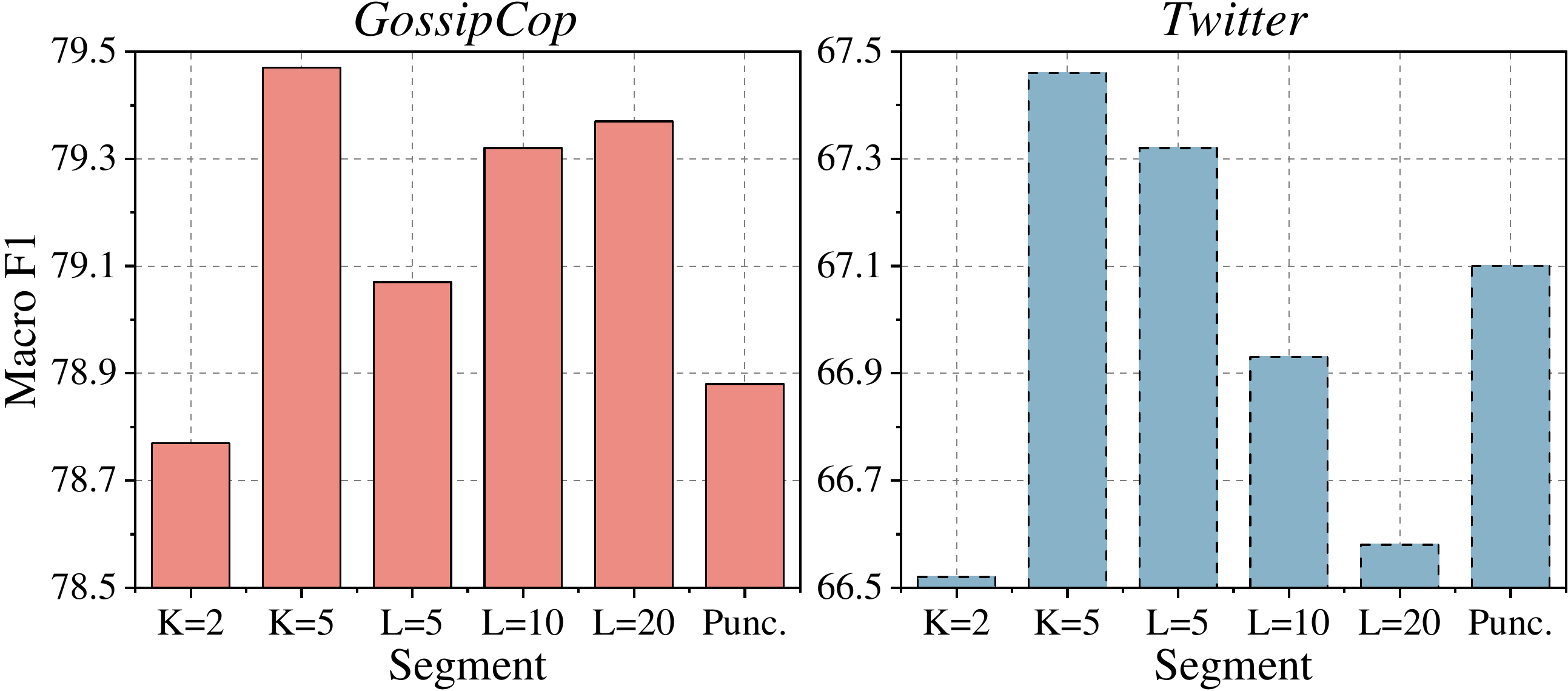}
  \caption{Sensitivity analysis of the segmentation strategies.}
  \label{sensitivity}
\end{figure}

\subsection{Sensitivity Analysis}

To select an appropriate text segmentation strategy for generating more accurate images that effectively replay the story conveyed in the text, we conduct a sensitivity analysis of various segmentation approaches. We experiment with three different segmentation strategies:
\begin{itemize}
    \item \textbf{Fixed-number segmentation}: The text is divided into a fixed number of $K$ segments, and $K$ images are generated to assist as augmentations. We select $K \in \{2, 5\}$ in the experiment;
    \item \textbf{Fixed-length segmentation}: The text is segmented into chunks of a fixed length $L$, and images are generated based on these segments. We select $L \in \{5, 10, 20\}$ in the experiment;
    \item \textbf{Punctuation-based segmentation}: The text is split at punctuation marks, but only when the segment contains more than 5 word tokens. These segments are then used for image generation.
\end{itemize}
The results of the sensitivity analysis are presented in Fig.~\ref{sensitivity}. We conduct experiments on the \textit{GossipCop} and \textit{Twitter} datasets, as both datasets consist of English-language content. Meanwhile, as shown in Table~\ref{datasetsta}, the average article lengths in \textit{GossipCop} and \textit{Twitter} are 566.2 and 16.5, respectively. This allows us to observe the performance of our method across different article-length scenarios.
Generally, the model achieves the best Macro F1 score when $K = 5$ and the worst performance when $K = 2$. This is because generating only two images renders our designed graph fusion strategy ineffective and makes it difficult to fully replay the story. When controlling for text length, the performance trends on the two datasets show opposite patterns: as text length increases, performance on the \textit{GossipCop} dataset consistently improves, while performance on the \textit{Twitter} dataset steadily declines. 
In the case of \textit{GossipCop}, when $L$ is relatively large (\eg $L = 10$ and $L = 20$), the performance is similar. However, when $L = 5$, the model generates an excessive number of images, significantly reducing efficiency while introducing some meaningless or redundant images, leading to a decline in performance. Conversely, on the \textit{Twitter} dataset, as $L$ increases, the number of text segments decreases significantly, sometimes resulting in only one image being generated, which also undermines the effectiveness of our graph fusion strategy.

\begin{figure}[t]
  \centering
  \includegraphics[scale=0.46]{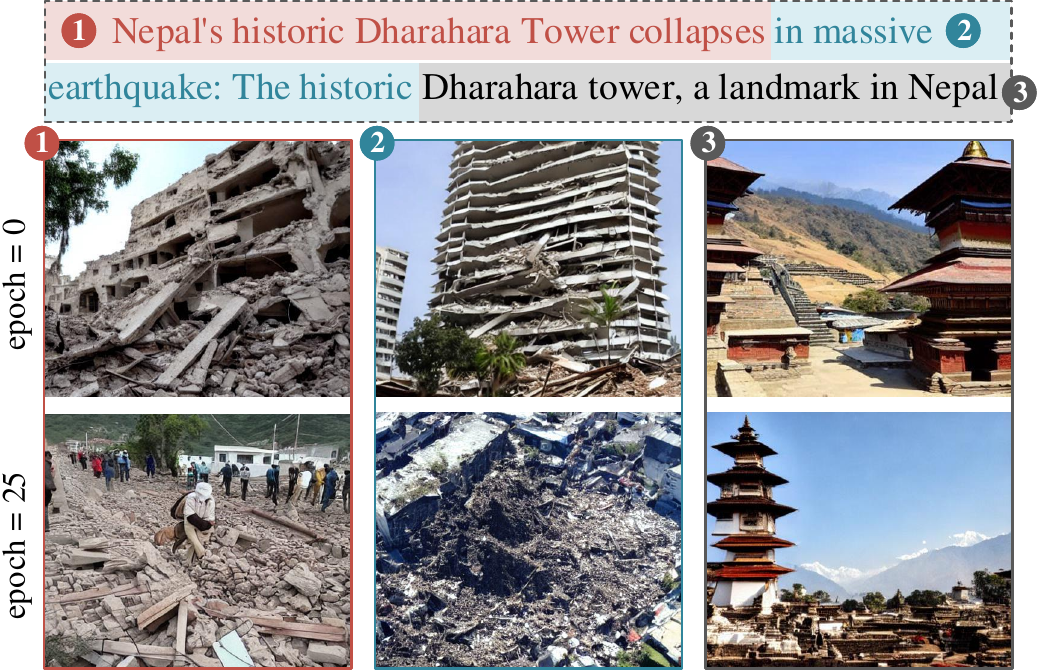}
  \caption{One representative sample is illustrated to demonstrate the performance of our text-to-image generator.}
  \label{casestudy}
\end{figure}

\subsection{Case Study}

We also present a representative example in Fig.~\ref{casestudy} to illustrate the comparison between images generated by an untrained generator and our trained generator for a given text. Generally, we first divide the text into $K = 3$ segments, each expressing different meanings, describing a collapsed tower, an earthquake, and a landmark tower, respectively. Both the untrained and trained generators are able to generate reasonable and high-quality images depicting these text segments.
In comparison, the images generated by the trained generator often contain more elements and present more information. For instance, regarding the collapsed tower, the trained generator produces images with more people, and for the earthquake, while the untrained generator generates an image of a collapsing building, the trained generator depicts a larger scene.

\subsection{Computational Budget}

In this section, we briefly present the GPU memory and time consumption for training our misinformation detector and text-to-image generator.
In experiments, training a misinformation detector requires approximately 1.2 hours with about 10GB GPU memory, and training a text-to-image generator takes around 5 hours with about 19GB GPU memory. Although training the generator takes longer than training the detector, our generator only needs to be trained once and can basically be applied to other detectors within the dataset.

\section{More Related Works}

In this section, we briefly review and organize the literature about text-to-image generation methods.
Over the past decade, Variational Auto-Encoders (VAE) \citep{kingma2014auto} and Generative Adversarial Networks (GAN) \citep{goodfellow2014generative}, along with their a series of variants \citep{oord2017neural,karras2019a,tao2022df}, have been the mainstream choice in image generation. Building upon them, DALL-E \citep{ramesh2021zero} is one of pioneer to incorporate the Transformer architecture \citep{vaswani2017attention} and utilize large-scale image-text pairs, \eg \textit{MS-COCO} \citep{lin2014microsoft}, for pre-training, achieving prominent results under zero-shot scenarios. 
However, some studies have found that these prevalent models suffer from issues such as training instability \citep{gulrajani2017improved,arjovsky2017wasserstein}. 
Consequently, recent diffusion models are proposed, which can be conditioned on various modality data and generate high-quality images. Among these, the most well-known is stable diffusion \citep{rombach2022high}, an open-source text-to-image generation model trained across the large-scale \textit{LAION} dataset \citep{schuhmann2022laion}. Our method employs this diffusion-based model as the image-text generator. 
Additionally, compared to similar image editing techniques \citep{avrahami2022blended,brooks2023instructpix2pix} in the community, text-to-image models generate images that are more diverse and flexible, rather than merely making partial editing on original images.

%% file: reference.bib
@inproceedings{devlin2019bert,
  author       = {Jacob Devlin and
                  Ming{-}Wei Chang and
                  Kenton Lee and
                  Kristina Toutanova},
  title        = {{BERT:} Pre-training of Deep Bidirectional Transformers for Language Understanding},
  booktitle    = {Conference of the North American Chapter of the Association for Computational Linguistics},
  pages        = {4171--4186},
  year         = {2019},
}

@inproceedings{chen2023causal,
  author       = {Ziwei Chen and
                  Linmei Hu and
                  Weixin Li and
                  Yingxia Shao and
                  Liqiang Nie},
  title        = {Causal Intervention and Counterfactual Reasoning for Multi-modal Fake News Detection},
  booktitle    = {Annual Meeting of the Association for Computational Linguistics},
  pages        = {627--638},
  year         = {2023},
}

@article{shu2020fakenewsnet,
  author       = {Kai Shu and
                  Deepak Mahudeswaran and
                  Suhang Wang and
                  Dongwon Lee and
                  Huan Liu},
  title        = {FakeNewsNet: {A} Data Repository with News Content, Social Context, and Spatiotemporal Information for Studying Fake News on Social Media},
  journal      = {Big Data},
  volume       = {8},
  number       = {3},
  pages        = {171--188},
  year         = {2020},
}

@inproceedings{jin2017multimodal,
  author       = {Zhiwei Jin and
                  Juan Cao and
                  Han Guo and
                  Yongdong Zhang and
                  Jiebo Luo},
  title        = {Multimodal Fusion with Recurrent Neural Networks for Rumor Detection on Microblogs},
  booktitle    = {{ACM} on Multimedia Conference},
  pages        = {795--816},
  year         = {2017},
}

@inproceedings{radford2021learning,
  author       = {Alec Radford and
                  Jong Wook Kim and
                  Chris Hallacy and
                  Aditya Ramesh and
                  Gabriel Goh and
                  Sandhini Agarwal and
                  Girish Sastry and
                  Amanda Askell and
                  Pamela Mishkin and
                  Jack Clark and
                  Gretchen Krueger and
                  Ilya Sutskever},
  title        = {Learning Transferable Visual Models From Natural Language Supervision},
  booktitle    = {International Conference on Machine Learning},
  volume       = {139},
  pages        = {8748--8763},
  year         = {2021},
}

@inproceedings{he2016deep,
  author       = {Kaiming He and
                  Xiangyu Zhang and
                  Shaoqing Ren and
                  Jian Sun},
  title        = {Deep Residual Learning for Image Recognition},
  booktitle    = {{IEEE} Conference on Computer Vision and Pattern Recognition},
  pages        = {770--778},
  year         = {2016},
}

@inproceedings{ying2023bootstrapping,
  author       = {Qichao Ying and
                  Xiaoxiao Hu and
                  Yangming Zhou and
                  Zhenxing Qian and
                  Dan Zeng and
                  Shiming Ge},
  title        = {Bootstrapping Multi-View Representations for Fake News Detection},
  booktitle    = {{AAAI} Conference on Artificial Intelligence},
  pages        = {5384--5392},
  year         = {2023},
}

@inproceedings{chen2022cross,
  author       = {Yixuan Chen and
                  Dongsheng Li and
                  Peng Zhang and
                  Jie Sui and
                  Qin Lv and
                  Tun Lu and
                  Li Shang},
  title        = {Cross-modal Ambiguity Learning for Multimodal Fake News Detection},
  booktitle    = {The {ACM} Web Conference},
  pages        = {2897--2905},
  year         = {2022},
}

@inproceedings{he2023debertav3,
  author       = {Pengcheng He and
                  Jianfeng Gao and
                  Weizhu Chen},
  title        = {DeBERTaV3: Improving DeBERTa using ELECTRA-Style Pre-Training with Gradient-Disentangled Embedding Sharing},
  booktitle    = {International Conference on Learning Representations},
  year         = {2023},
}

@inproceedings{he2022masked,
  author       = {Kaiming He and
                  Xinlei Chen and
                  Saining Xie and
                  Yanghao Li and
                  Piotr Doll{\'{a}}r and
                  Ross B. Girshick},
  title        = {Masked Autoencoders Are Scalable Vision Learners},
  booktitle    = {{IEEE/CVF} Conference on Computer Vision and Pattern Recognition},
  pages        = {15979--15988},
  year         = {2022},
}

@article{wang2024escaping,
  author       = {Bing Wang and
                  Ximing Li and
                  Changchun Li and
                  Shengsheng Wang and
                  Wanfu Gao},
  title        = {Escaping the Neutralization Effect of Modality Features Fusion in Multimodal Fake News Detection},
  journal      = {Information Fusion},
  volume       = {111},
  pages        = {102500},
  year         = {2024},
}

@inproceedings{wang2024harmfully,
  author       = {Bing Wang and
                  Sheng{-}Sheng Wang and
                  Changchun Li and
                  Renchu Guan and
                  Ximing Li},
  title        = {Harmfully Manipulated Images Matter in Multimodal Misinformation Detection},
  booktitle    = {{ACM} International Conference on Multimedia},
  pages        = {2262--2271},
  year         = {2024},
}

@inproceedings{wang2024why,
  author       = {Bing Wang and
                  Ximing Li and
                  Changchun Li and
                  Bo Fu and
                  Songwen Pei and
                  Shengsheng Wang},
  title        = {Why Misinformation is Created? Detecting them by Integrating Intent Features},
  booktitle    = {{ACM} International Conference on Information and Knowledge Management},
  pages        = {2304--2314},
  year         = {2024},
}

@article{boididou2018detection,
  author       = {Christina Boididou and
                  Symeon Papadopoulos and
                  Markos Zampoglou and
                  Lazaros Apostolidis and
                  Olga Papadopoulou and
                  Yiannis Kompatsiaris},
  title        = {Detection and visualization of misleading content on Twitter},
  journal      = {International Journal of Multimedia Information Retrieval},
  volume       = {7},
  number       = {1},
  pages        = {71--86},
  year         = {2018},
}

@article{quinlan1986induction,
  author       = {J. Ross Quinlan},
  title        = {Induction of Decision Trees},
  journal      = {Machine Learning},
  volume       = {1},
  number       = {1},
  pages        = {81--106},
  year         = {1986},
}

@article{donoho1995de,
  author       = {David L. Donoho},
  title        = {De-noising by soft-thresholding},
  journal      = {IEEE Transactions on Information Theory},
  volume       = {41},
  number       = {3},
  pages        = {613--627},
  year         = {1995},
}

@inproceedings{schuhmann2022laion,
  author       = {Christoph Schuhmann and
                  Romain Beaumont and
                  Richard Vencu and
                  Cade Gordon and
                  Ross Wightman and
                  Mehdi Cherti and
                  others},
  title        = {{LAION-5B:} An open large-scale dataset for training next generation image-text models},
  booktitle    = {Advances in Neural Information Processing Systems},
  year         = {2022},
}

@inproceedings{rombach2022high,
  author       = {Robin Rombach and
                  Andreas Blattmann and
                  Dominik Lorenz and
                  Patrick Esser and
                  Bj{\"{o}}rn Ommer},
  title        = {High-Resolution Image Synthesis with Latent Diffusion Models},
  booktitle    = {{IEEE/CVF} Conference on Computer Vision and Pattern Recognition},
  pages        = {10674--10685},
  year         = {2022},
}

@inproceedings{hu2022lora,
  author       = {Edward J. Hu and
                  Yelong Shen and
                  Phillip Wallis and
                  Zeyuan Allen{-}Zhu and
                  Yuanzhi Li and
                  Shean Wang and
                  Lu Wang and
                  Weizhu Chen},
  title        = {LoRA: Low-Rank Adaptation of Large Language Models},
  booktitle    = {International Conference on Learning Representations},
  year         = {2022},
}

@inproceedings{wu2021multimodal,
  author       = {Yang Wu and
                  Pengwei Zhan and
                  Yunjian Zhang and
                  LiMing Wang and
                  Zhen Xu},
  title        = {Multimodal Fusion with Co-Attention Networks for Fake News Detection},
  booktitle    = {Findings of the Association for Computational Linguistics: {ACL/IJCNLP}},
  volume       = {{ACL/IJCNLP} 2021},
  pages        = {2560--2569},
  year         = {2021},
}

@inproceedings{zhou2020safe,
  author       = {Xinyi Zhou and
                  Jindi Wu and
                  Reza Zafarani},
  title        = {{SAFE:} Similarity-Aware Multi-modal Fake News Detection},
  booktitle    = {Advances in Knowledge Discovery and Data Mining - 24th Pacific-Asia Conference},
  volume       = {12085},
  pages        = {354--367},
  year         = {2020},
}

@inproceedings{krizhevsky2012imagenet,
  author       = {Alex Krizhevsky and
                  Ilya Sutskever and
                  Geoffrey E. Hinton},
  title        = {ImageNet Classification with Deep Convolutional Neural Networks},
  booktitle    = {Advances in Neural Information Processing Systems},
  pages        = {1106--1114},
  year         = {2012},
}

@inproceedings{ma2024event,
  author       = {Zihan Ma and
                  Minnan Luo and
                  Hao Guo and
                  Zhi Zeng and
                  Yiran Hao and
                  Xiang Zhao},
  title        = {Event-Radar: Event-driven Multi-View Learning for Multimodal Fake News Detection},
  booktitle    = {Annual Meeting of the Association for Computational Linguistics},
  pages        = {5809--5821},
  year         = {2024},
}

@inproceedings{wang2024fake,
  author       = {Jiandong Wang and
                  Hongguang Zhang and
                  Chun Liu and
                  Xiongjun Yang},
  title        = {Fake News Detection via Multi-scale Semantic Alignment and Cross-modal Attention},
  booktitle    = {International {ACM} {SIGIR} Conference on Research and Development in Information Retrieval},
  pages        = {2406--2410},
  year         = {2024},
}

@article{xuan2024lemma,
  author       = {Keyang Xuan and
                  Li Yi and
                  Fan Yang and
                  Ruochen Wu and
                  Yi R. Fung and
                  Heng Ji},
  title        = {{LEMMA:} Towards LVLM-Enhanced Multimodal Misinformation Detection with External Knowledge Augmentation},
  journal      = {CoRR},
  volume       = {abs/2402.11943},
  year         = {2024},
}

@inproceedings{fung2021infosurgeon,
  author       = {Yi R. Fung and
                  Christopher Thomas and
                  Revanth Gangi Reddy and
                  Sandeep Polisetty and
                  Heng Ji and
                  Shih{-}Fu Chang and
                  Kathleen R. McKeown and
                  Mohit Bansal and
                  Avi Sil},
  title        = {InfoSurgeon: Cross-Media Fine-grained Information Consistency Checking for Fake News Detection},
  booktitle    = {Annual Meeting of the Association for Computational Linguistics},
  pages        = {1683--1698},
  year         = {2021},
}

@inproceedings{dong2024unveiling,
  author       = {Yiqi Dong and
                  Dongxiao He and
                  Xiaobao Wang and
                  Youzhu Jin and
                  Meng Ge and
                  Carl Yang and
                  Di Jin},
  title        = {Unveiling Implicit Deceptive Patterns in Multi-Modal Fake News via Neuro-Symbolic Reasoning},
  booktitle    = {{AAAI} Conference on Artificial Intelligence},
  pages        = {8354--8362},
  year         = {2024},
}

@inproceedings{wu2023see,
  author       = {Lianwei Wu and
                  Pusheng Liu and
                  Yanning Zhang},
  title        = {See How You Read? Multi-Reading Habits Fusion Reasoning for Multi-Modal Fake News Detection},
  booktitle    = {{AAAI} Conference on Artificial Intelligence},
  pages        = {13736--13744},
  year         = {2023},
}

@article{hu2023causal,
  author       = {Linmei Hu and
                  Ziwei Chen and
                  Ziwang Zhao and
                  Jianhua Yin and
                  Liqiang Nie},
  title        = {Causal Inference for Leveraging Image-Text Matching Bias in Multi-Modal Fake News Detection},
  journal      = {{IEEE} Transactions on Knowledge and Data Engineering},
  volume       = {35},
  number       = {11},
  pages        = {11141--11152},
  year         = {2023},
}

@inproceedings{ramesh2021zero,
  author       = {Aditya Ramesh and
                  Mikhail Pavlov and
                  Gabriel Goh and
                  Scott Gray and
                  Chelsea Voss and
                  Alec Radford and
                  Mark Chen and
                  Ilya Sutskever},
  title        = {Zero-Shot Text-to-Image Generation},
  booktitle    = {International Conference on Machine Learning},
  volume       = {139},
  pages        = {8821--8831},
  year         = {2021},
}

@inproceedings{kingma2014auto,
  author       = {Diederik P. Kingma and
                  Max Welling},
  title        = {Auto-Encoding Variational Bayes},
  booktitle    = {International Conference on Learning Representations},
  year         = {2014},
}

@inproceedings{goodfellow2014generative,
  author       = {Ian J. Goodfellow and
                  Jean Pouget{-}Abadie and
                  Mehdi Mirza and
                  Bing Xu and
                  David Warde{-}Farley and
                  Sherjil Ozair and
                  Aaron C. Courville and
                  Yoshua Bengio},
  title        = {Generative Adversarial Nets},
  booktitle    = {Advances in Neural Information Processing Systems},
  pages        = {2672--2680},
  year         = {2014},
}

@inproceedings{oord2017neural,
  author       = {A{\"{a}}ron van den Oord and
                  Oriol Vinyals and
                  Koray Kavukcuoglu},
  title        = {Neural Discrete Representation Learning},
  booktitle    = {Advances in Neural Information Processing Systems},
  pages        = {6306--6315},
  year         = {2017},
}

@inproceedings{tao2022df,
  author       = {Ming Tao and
                  Hao Tang and
                  Fei Wu and
                  Xiaoyuan Jing and
                  Bing{-}Kun Bao and
                  Changsheng Xu},
  title        = {{DF-GAN:} {A} Simple and Effective Baseline for Text-to-Image Synthesis},
  booktitle    = {{IEEE/CVF} Conference on Computer Vision and Pattern Recognition},
  pages        = {16494--16504},
  year         = {2022},
}

@inproceedings{karras2019a,
  author       = {Tero Karras and
                  Samuli Laine and
                  Timo Aila},
  title        = {A Style-Based Generator Architecture for Generative Adversarial Networks},
  booktitle    = {{IEEE} Conference on Computer Vision and Pattern Recognition},
  pages        = {4401--4410},
  year         = {2019},
}

@inproceedings{vaswani2017attention,
  author       = {Ashish Vaswani and
                  Noam Shazeer and
                  Niki Parmar and
                  Jakob Uszkoreit and
                  Llion Jones and
                  Aidan N. Gomez and
                  Lukasz Kaiser and
                  Illia Polosukhin},
  title        = {Attention is All you Need},
  booktitle    = {Advances in Neural Information Processing Systems},
  pages        = {5998--6008},
  year         = {2017},
}

@inproceedings{lin2014microsoft,
  author       = {Tsung{-}Yi Lin and
                  Michael Maire and
                  Serge J. Belongie and
                  James Hays and
                  Pietro Perona and
                  Deva Ramanan and
                  Piotr Doll{\'{a}}r and
                  C. Lawrence Zitnick},
  title        = {Microsoft {COCO:} Common Objects in Context},
  booktitle    = {European Conference on Computer Vision},
  volume       = {8693},
  pages        = {740--755},
  year         = {2014},
}

@inproceedings{gulrajani2017improved,
  author       = {Ishaan Gulrajani and
                  Faruk Ahmed and
                  Mart{\'{\i}}n Arjovsky and
                  Vincent Dumoulin and
                  Aaron C. Courville},
  title        = {Improved Training of Wasserstein GANs},
  booktitle    = {Advances in Neural Information Processing Systems},
  pages        = {5767--5777},
  year         = {2017},
}

@inproceedings{arjovsky2017wasserstein,
  author       = {Mart{\'{\i}}n Arjovsky and
                  Soumith Chintala and
                  L{\'{e}}on Bottou},
  title        = {Wasserstein Generative Adversarial Networks},
  booktitle    = {International Conference on Machine Learning},
  volume       = {70},
  pages        = {214--223},
  year         = {2017},
}

@inproceedings{brooks2023instructpix2pix,
  author       = {Tim Brooks and
                  Aleksander Holynski and
                  Alexei A. Efros},
  title        = {InstructPix2Pix: Learning to Follow Image Editing Instructions},
  booktitle    = {{IEEE/CVF} Conference on Computer Vision and Pattern Recognition},
  pages        = {18392--18402},
  year         = {2023},
}

@inproceedings{avrahami2022blended,
  author       = {Omri Avrahami and
                  Dani Lischinski and
                  Ohad Fried},
  title        = {Blended Diffusion for Text-driven Editing of Natural Images},
  booktitle    = {{IEEE/CVF} Conference on Computer Vision and Pattern Recognition},
  pages        = {18187--18197},
  year         = {2022},
}

@inproceedings{zhang2021mining,
  author       = {Xueyao Zhang and
                  Juan Cao and
                  Xirong Li and
                  Qiang Sheng and
                  Lei Zhong and
                  Kai Shu},
  title        = {Mining Dual Emotion for Fake News Detection},
  booktitle    = {The Web Conference},
  pages        = {3465--3476},
  year         = {2021},
}

@inproceedings{zhu2022generalizing,
  author       = {Yongchun Zhu and
                  Qiang Sheng and
                  Juan Cao and
                  Shuokai Li and
                  Danding Wang and
                  Fuzhen Zhuang},
  title        = {Generalizing to the Future: Mitigating Entity Bias in Fake News Detection},
  booktitle    = {International {ACM} {SIGIR} Conference on Research and Development in Information Retrieval},
  pages        = {2120--2125},
  year         = {2022},
}

@inproceedings{kipf2017semi,
  author       = {Thomas N. Kipf and
                  Max Welling},
  title        = {Semi-Supervised Classification with Graph Convolutional Networks},
  booktitle    = {International Conference on Learning Representations},
  year         = {2017},
}

@inproceedings{clark2023text,
  author       = {Kevin Clark and
                  Priyank Jaini},
  title        = {Text-to-Image Diffusion Models are Zero Shot Classifiers},
  booktitle    = {Advances in Neural Information Processing Systems},
  year         = {2023},
}

@inproceedings{zhang2021cross,
  author       = {Han Zhang and
                  Jing Yu Koh and
                  Jason Baldridge and
                  Honglak Lee and
                  Yinfei Yang},
  title        = {Cross-Modal Contrastive Learning for Text-to-Image Generation},
  booktitle    = {{IEEE} Conference on Computer Vision and Pattern Recognition},
  pages        = {833--842},
  year         = {2021},
}

@article{vosoughi2018spread,
  title={The spread of true and false news online},
  author={Vosoughi, Soroush and Roy, Deb and Aral, Sinan},
  journal={science},
  volume={359},
  number={6380},
  pages={1146--1151},
  year={2018},
}

@article{scheufele2019science,
  title={Science audiences, misinformation, and fake news},
  author={Scheufele, Dietram A and Krause, Nicole M},
  journal={Proceedings of the National Academy of Sciences},
  volume={116},
  number={16},
  pages={7662--7669},
  year={2019},
}

@inproceedings{zeng2024multimodal,
  author       = {Fengzhu Zeng and
                  Wenqian Li and
                  Wei Gao and
                  Yan Pang},
  title        = {Multimodal Misinformation Detection by Learning from Synthetic Data with Multimodal LLMs},
  booktitle    = {Findings of the Association for Computational Linguistics: {EMNLP}},
  pages        = {10467--10484},
  year         = {2024},
}

@inproceedings{tahmasebi2024multimodal,
  author       = {Sahar Tahmasebi and
                  Eric M{\"{u}}ller{-}Budack and
                  Ralph Ewerth},
  title        = {Multimodal Misinformation Detection using Large Vision-Language Models},
  booktitle    = {{ACM} International Conference on Information and Knowledge Management},
  pages        = {2189--2199},
  year         = {2024},
}

@inproceedings{liu2024fka,
  author       = {Xuannan Liu and
                  Peipei Li and
                  Huaibo Huang and
                  Zekun Li and
                  Xing Cui and
                  Jiahao Liang and
                  Lixiong Qin and
                  Weihong Deng and
                  Zhaofeng He},
  title        = {FKA-Owl: Advancing Multimodal Fake News Detection through Knowledge-Augmented LVLMs},
  booktitle    = {{ACM} International Conference on Multimedia},
  pages        = {10154--10163},
  year         = {2024},
}

@inproceedings{shen2025gamed,
  author       = {Lingzhi Shen and
                  Yunfei Long and
                  Xiaohao Cai and
                  Imran Razzak and
                  Guanming Chen and
                  Kang Liu and
                  Shoaib Jameel},
  title        = {{GAMED:} Knowledge Adaptive Multi-Experts Decoupling for Multimodal Fake News Detection},
  booktitle    = {{ACM} International Conference on Web Search and Data Mining},
  pages        = {586--595},
  year         = {2025},
}

@article{mittal2013making,
  author       = {Anish Mittal and
                  Rajiv Soundararajan and
                  Alan C. Bovik},
  title        = {Making a "Completely Blind" Image Quality Analyzer},
  journal      = {{IEEE} Signal Processing Letters},
  volume       = {20},
  number       = {3},
  pages        = {209--212},
  year         = {2013},
}

@inproceedings{n2015blind,
  author       = {Venkatanath N. and
                  Praneeth D. and
                  Maruthi Chandrasekhar Bh. and
                  Sumohana S. Channappayya and
                  Swarup S. Medasani},
  title        = {Blind image quality evaluation using perception based features},
  booktitle    = {National Conference on Communications},
  pages        = {1--6},
  year         = {2015},
}
